\documentclass[a4paper,fleqn]{cas-dc}

\usepackage[numbers]{natbib}
\usepackage{bm}
\usepackage{subcaption}
\usepackage{eso-pic}

\def\tsc#1{\csdef{#1}{\textsc{\lowercase{#1}}\xspace}}
\tsc{WGM}
\tsc{QE}

\AddToShipoutPictureBG*{%
  \AtPageUpperLeft{%
    \hspace{2cm}%
    \raisebox{-14pt}{%
      \makebox[0pt][l]{%
        \begin{minipage}{15cm}
          \fontsize{7}{8}\selectfont This article has been accepted for publication in Computers in Biology and Medicine (Elsevier). This is the author's accepted version of the manuscript which has not been fully edited and content may change prior to final publication. Citation information: DOI 10.1016/j.compbiomed.2026.111809
        \end{minipage}
      }
    }
  }
  \AtPageLowerLeft{%
    \hspace{3.6cm}%
    \raisebox{10pt}{%
      \makebox[0pt][l]{%
        \begin{minipage}{15cm}
          \fontsize{7}{8}\selectfont ©2026. This manuscript version is made available under the CC-BY-NC-ND 4.0 license (\url{https://creativecommons.org/licenses/by-nc-nd/4.0}).
        \end{minipage}
      }
    }
  }
}

\begin{document}
\let\WriteBookmarks\relax
\def\floatpagepagefraction{1}
\def\textpagefraction{.001}

\shorttitle{}    

\shortauthors{}  

\title [mode = title]{Evaluating Interactive 2D Visualization as a Sample Selection Strategy for Biomedical Time‑Series Data Annotation}


\author[1]{Einari Vaaras}[type=editor,
                      auid=000,
                      bioid=1,
                      orcid=0000-0002-8714-6090]
\cormark[1]
\ead{einari.vaaras@tuni.fi}
\credit{Conceptualization, Methodology, Software, Investigation, Data Curation, Writing – original draft, Visualization}

\affiliation[1]{organization={Signal Processing Research Centre, Tampere University},
                city={Tampere},
                postcode={33720}, 
                country={Finland}}

\author[2]{Manu Airaksinen}[auid=001,
                            bioid=2,
                            orcid=0000-0002-8031-2260]
\credit{Conceptualization, Writing - Review \& Editing, Supervision}

\author[1]{Okko Räsänen}[auid=002,
                        bioid=3,
                        orcid=0000-0002-0537-0946]
\credit{Conceptualization, Writing - Review \& Editing, Supervision}

\affiliation[2]{organization={BABA Center, Department of Physiology, University of Helsinki},
                city={Helsinki},
                postcode={00029}, 
                country={Finland}}

\cortext[cor1]{Corresponding author}

\begin{abstract}
Reliable machine-learning models in biomedical settings depend on accurate labels, yet annotating biomedical time-series data remains challenging. Algorithmic sample selection may support annotation, but evidence from studies involving real human annotators is scarce. Consequently, we compare three sample selection methods for annotation: random sampling (RND), farthest‑first traversal (FAFT), and a graphical user interface-based method enabling exploration of complementary 2D visualizations (2DVs) of high‑dimensional data. We evaluated the methods across four classification tasks in infant motility assessment (IMA) and speech emotion recognition (SER). Twelve annotators, categorized as experts or non‑experts, performed data annotation under a limited annotation budget, and post-annotation experiments were conducted to evaluate the sampling methods. Across all classification tasks, 2DV performed best when aggregating labels across annotators. In IMA, 2DV most effectively captured rare classes, but also exhibited greater annotator-to-annotator label distribution variability resulting from the limited annotation budget, decreasing classification performance when models were trained on individual annotators' labels; in these cases, FAFT excelled. For SER, 2DV outperformed the other methods among expert annotators and matched their performance for non‑experts in the individual-annotator setting. A failure risk analysis revealed that RND was the safest choice when annotator count or annotator expertise was uncertain, whereas 2DV had the highest risk due to its greater label distribution variability. Furthermore, post-experiment interviews indicated that 2DV made the annotation task more interesting and enjoyable. Overall, 2DV-based sampling appears promising for biomedical time-series data annotation, particularly when the annotation budget is not highly constrained. The annotation software is freely available at \url{https://github.com/SPEECHCOG/TSExplorer}.


\end{abstract}




\maketitle

\section{Introduction} \label{sec_introduction}

Annotating biomedical time‑series data, often consisting of physiological signals (e.g. electroencephalography (EEG), electromyography (EMG), or actigraphy) and/or highly specific classification tasks (e.g. polysomnography (PSG)-based sleep stage classification \cite{psg_al_initialization}), is frequently inconsistent and time-consuming \cite{watson_clinical_data_annotation_time_consuming, impact_of_inconsistent_annotations_ai_driven_decision_making, interrator_reliability_estimators, eeg_improved_manual_annotation, multi_interactive_learning_model_sleep_staging, robustsleepnet}. Yet, high‑quality labels are essential for training reliable machine‑learning (ML) models that are required in healthcare settings \cite{impact_of_inconsistent_annotations_ai_driven_decision_making, iar_2_0, medical_image_analysis_noise_labels_survey, dl_noisy_labels_medical_prediction_problems, noise_in_gold_labels, deep_self_cleansing_with_noisy_labels, benchmarking_real_world_medical_image_survey_paper}. Both sample selection and active learning (AL) methods have been proposed to optimize the annotation process by performing algorithm-driven intelligent sample selection with the formulated goal of ``maximal performance with minimal annotations'' (see Section \ref{sec_related_work}). However, the sample selection in biomedical classification tasks may be difficult to outsource to an algorithm. Further, almost all prior work in sample selection and AL is based on simulations where the researchers use annotations from a previously annotated dataset (where the given sample selection strategy was not used) for their experiments, which assumes statistical independence between the annotator and the process in which the annotator is engaged in. This does not take into account intra-annotator effects that might be caused, for example, by cognitive fatigue due to overly monotonous or excessively high sample variability.

One potential and often overlooked approach for supporting efficient and interpretable data annotation workflows is the use of interactive 2D visualizations (2DVs), which have been shown to be beneficial across a wide range of contexts. For example, they can increase the efficiency of data annotation \cite{vision6d, umap_imaging_flow_cytometry, fallgren_interspeech}, enhance interpretability of high-dimensional data \cite{vision6d, t_visne_article, voxplorer, CZ_CELLxGENE, cytometry_2d_embeddings, dino_clinical_data_2d}, and reveal patterns that are difficult to detect using conventional analysis methods \cite{visne_article, cytometry_2d_embeddings}. While prior work (Section \ref{sec_related_work}) has demonstrated the utility of 2DV for data analysis and annotation, none of these studies have examined how annotation choices influence label distributions or downstream classification performance, both of which are factors that are central to the practical applicability of ML models. Moreover, existing approaches have not evaluated multiple data modalities or classification tasks, nor have they investigated annotation under a fixed labeling budget in scenarios where only a small proportion of the available data can be labeled.

To address these limitations, we conduct a proof-of-concept study that makes two main contributions: First, we present the Time-Series Explorer (TSExplorer), a general-purpose graphical user interface (GUI) framework for time-series data annotation and exploration. TSExplorer visualizes the entire dataset as a 2D scatter plot and allows annotators to freely explore complementary 2D representations of the underlying high-dimensional data. Second, we empirically compare three sample selection methods, namely random sampling (RND), farthest-first traversal (FAFT), and a GUI-based approach utilizing 2DVs with the TSExplorer software, using real human annotators. We evaluate the three sample selection methods across two biomedical datasets spanning four classification tasks. Specifically, our study examines infant posture and movement classification within infant motility assessment (IMA) using multi-sensor inertial measurement unit (IMU) recordings, as well as valence and arousal classification within speech emotion recognition (SER) in a neonatal intensive care unit (NICU) setting.

Our proof-of-concept experiments involved 12 human annotators altogether, and annotators were categorized as either experts or non-experts. In the context of the present study, \textit{experts} were defined as individuals with previous annotation experience relevant to the given data domain: IMA annotation experience for multi-sensor IMU data, or SER annotation experience for child-centered audio recordings. Annotators without such prior experience were categorized as \textit{non-experts}. After the annotation process, we conducted a set of post-annotation experiments to reveal the trade-offs between the sample selection methods.

This paper is organized as follows: First, Section \ref{sec_related_work} provides an overview of research relevant to this study, followed by a description of the used sample selection methods in Section \ref{sec_sample_selection_methods}. Next, the experimental setup and its details are presented in Section \ref{sec_experimental_setup}, and the results of the experiments are then provided in Section \ref{sec_results}. Finally, Section \ref{sec_conclusion} summarizes the conclusions, discusses the limitations of the present experiments, and highlights potential directions for future work.

\section{Related work} \label{sec_related_work}

Although the present study focuses on sample selection rather than AL, the two concepts are closely related as both aim to prioritize informative samples under limited annotation resources: AL methods typically rely on iterative feedback loops between the ground truth information source (e.g. a human annotator) and the sample selection algorithm, whereas sample selection in our setting is performed without such interaction. Nevertheless, prior AL literature is directly relevant to this study, as it provides algorithmic strategies for identifying samples that improve model performance while reducing labeling effort. In the context of biomedical time-series data, AL and sample selection methods have been explored in multiple domains. For example, Bi et al. \cite{dynamic_al_imu_har} proposed a dynamic AL approach for IMU-based human activity recognition (HAR) which simultaneously selects informative samples and identifies previously unseen activity categories. Their results demonstrated that the proposed approach reduced annotation workload, particularly in cases with a limited annotation budget. Similarly, Ding et al. \cite{al_emg_har} introduced an AL framework for EMG-based HAR that selects informative samples through data screening and similarity modeling, demonstrating improved cross-subject model generalization. Furthermore, Siuly et al. \cite{eeg_sample_selection} compared two sampling strategies for EEG signals, and showed that random sampling provided more representative EEG samples and led to superior downstream classification performance. Grimova et al. \cite{psg_al_initialization} introduced an initialization strategy for AL in PSG-based sleep stage classification, and demonstrated that their method alleviates poor early-stage performance issues compared to using $k$-means clustering-based initialization. Li et al. \cite{al_semisupervised_ecg} presented an AL framework for electrocardiogram (ECG) analysis that automatically identifies informative samples for semi-supervised learning. Their framework enhanced performance on both segmentation and classification tasks while reducing the amount of labeled data required. Holgado-Cuadrado et al. \cite{al_long_term_ecg} developed an AL approach for ECG analysis that selects informative samples using margin sampling in an autoencoder‑based latent space. They showed that their method can be used to reduce annotation workload, and that the method consistently outperformed random sampling across three classification tasks.

The vast majority of studies regarding algorithmic sample selection rely on simulated annotation procedures, where labels from previously annotated datasets are reused to simulate annotator decisions. Only a limited number of algorithmic sample selection studies have incorporated real human annotators into the annotation process. For instance, Settles et al. \cite{settles_al_with_real_annotation_costs} studied various aspects of real-life annotation cost in AL across four datasets from image and text domains, demonstrating notable variability in annotation costs between experimental conditions. Baldridge and Palmer \cite{al_two_annotators_different_expertise} investigated AL using two annotators with different background expertise, and found that the effectiveness of any given annotation strategy depended on annotator expertise. Liu et al. \cite{al_human_pose_estimation}, Wei et al. \cite{cost_aware_al_clinical_text}, and Lee et al. \cite{annotation_curricula} conducted their main sample selection experiments using a simulated annotation procedure, but validated their approaches using human annotators. Calma et al. \cite{al_with_realistic_data_case_study} conducted a study using two human annotators and confirmed that their AL approach performed as expected in practice. Vaaras et al. \cite{nicu_a_speechcomm} used a clustering-based AL algorithm and Lahtinen et al. \cite{finnaffect_speechcomm} used the CLARA clustering algorithm to select a subset from a large pool of unlabeled samples for data annotation when constructing annotated corpora.

FAFT is a sampling strategy traditionally used for initializing clustering algorithms, and it has also been widely applied in AL (see e.g. \cite{al_text_datasets, mal_paper, al_cnn_coreset, al_sed}) due to its strong geometric coverage of the feature space \cite{faft_original_gonzalez, clustering_without_over_representation}. FAFT-based data sampling has been previously used in SER on multiple occasions \cite{nicu_a_speechcomm, vaaras22_interspeech, lahtinen25_interspeech}. However, to the best of our knowledge, FAFT has not been previously used in HAR or other related biomedical time-series data domains (such as EEG or EMG), with the exception of the study by Zhang et al. \cite{faft_al_ecg}, where FAFT was used for sample selection in AL for ECG data. On the other hand, other sample selection strategies have been examined (using simulated experiments) in IMU-based HAR. Alharbi et al. \cite{sampling_strategy_comparison_har_imbalanced_data} proposed three sample selection strategies to tackle class imbalances, and showed through experiments with benchmark datasets that their sampling strategies reduced this effect. Wei et al. \cite{activeselfhar} introduced a framework that combines model confidence and boundary‑focused querying to identify informative samples for cross‑subject adaptation, demonstrating that carefully selected target domain samples can substantially reduce labeling effort without degrading HAR performance. Moreover, several HAR survey papers (e.g. \cite{HAR_tutorial, recent_trends_in_ml_for_har_survey, leveraging_al_in_har, transfer_learning_har_survey}) emphasize that data sampling and sample selection strategies, even though not widely explored, are increasingly necessary in practical HAR applications.

A separate line of related work concerns human-in-the-loop data exploration using 2DVs, with earlier work spanning multiple domains and providing tools that support data interpretation, interactive analysis, or annotation depending on the application context. However, applications to biomedical time‑series data annotation remain limited. One example is the 3D‑to‑2D interactive visualization and annotation tool by Zhang et al. \cite{vision6d} which enables users to explore 6D pose estimation data through synchronized 3D views and corresponding 2D projections. In the paper, the authors demonstrated that the possibility of visualizing and manipulating 3D objects interactively on 2D real-world scenes made it easier and more efficient to annotate and refine 6D poses. Another example is the interactive GUI by Chatzimparmpas et al. \cite{t_visne_article}, which was designed for better understanding t-distributed stochastic neighbor embedding (t-SNE) projections through visual investigation of a collection of different projections. Specifically, their GUI allows users to explore the effect of different hyperparameter values on t-SNE, interact with the scatter plots e.g. through investigating correlations between dimensions and visual patterns, and assess the quality of visualizations e.g. via inspecting sample densities or neighborhood preservation. The authors demonstrated the usefulness of their GUI through hypothetical use cases with real-world data and a user study with human participants, highlighting how interactive visualization can support data interpretability.

Interactive 2D data exploration has also been used in analyzing audio and large‑scale biological data. Voxplorer \cite{voxplorer} presents an interactive dashboard for exploring and visualizing voice data using either 2D or 3D embeddings from dimensionality reduction algorithms such as t-SNE and uniform manifold approximation and projection (UMAP). Combined with the possibility of extracting features using the dashboard, Voxplorer aims to give researchers access to modern data analysis techniques. The CZ CELLxGENE Discover platform \cite{CZ_CELLxGENE} serves as a large-scale, curated repository of community-contributed single-cell transcriptomic data. In addition to enabling access to corpus‑level analyses, the platform provides an interactive 2DV interface that allows researchers to explore and analyze individual datasets using 2D embeddings from algorithms such as t‑SNE, UMAP, and principal component analysis (PCA).

Tools for 2D data visualization have additionally been evaluated in biomedical contexts, most commonly in cytometry. Amir et al. \cite{visne_article} introduced an application of t-SNE to cytometry, demonstrating that 2D embeddings can improve the visibility of leukemic cell pattern differences that are difficult to detect through manual gating. Park et al. \cite{cytometry_2d_embeddings} evaluated how well t-SNE and UMAP dimensionality reduction algorithms can support diagnostic decision‑making in flow cytometry. They showed that the results from these algorithms closely match the results obtained through traditional manual analysis, and in some cases even highlight small or unusual cell groups that the manual approach did not detect. Stolarek et al. \cite{umap_imaging_flow_cytometry} evaluated UMAP for exploring high‑dimensional data in imaging flow cytometry, showing that the use of 2D embeddings from UMAP improves clustering quality and supports rapid grouping and provisional labeling of previously unannotated cells.

In biomedical contexts outside cytometry, 2D data visualization has also shown potential. Yang et al. \cite{clinical_data_2d_visualization_comparison} compared different 2DV algorithms, and found that among the tested algorithms UMAP most effectively separated biologically and clinically relevant groups in population-level gene expression data. Zhou et al. \cite{spasne} proposed a dimensionality reduction method specifically tailored for integrating both spatial and molecular information into low-dimensional embeddings. This, in turn, enabled more accurate visualization of tissue organization across diseased and normal samples in spatial gene expression datasets. Similarly, Hoq et al. \cite{dino_clinical_data_2d} investigated the use of 2D embeddings obtained from a self-supervised foundation model pre-trained using X-ray medical images. In the paper, the authors demonstrated that such embeddings can help to distinguish cancer-associated patterns from non-cancer cases in lung cancer screening data.

Conceptually the closest predecessor to our work is the annotation GUI proposed by Fallgren et al. \cite{fallgren_interspeech}. Their approach presents a highly efficient workflow for annotating large‑scale audio data through the use of precomputed embeddings, 2D visual projections, and interactive cluster exploration. In their system, audio segments are mapped into 2D space using dimensionality reduction algorithms such as t-SNE and UMAP, with each data point corresponding to an audio segment, and annotators explore this space by listening to either a single audio segment or a blend of multiple segments at a time. Annotators could change the 2DV algorithm at any time, and samples could be annotated in bulk by painting a set of points with a color corresponding to a specific label. Their approach also incorporated data sampling by visualizing a downsampled view of the data points from which labels are then propagated to the full set of points using nearest-neighbor propagation, further emphasizing that their GUI served as a tool for bulk labeling. In their experiments with speech activity detection, the authors demonstrated that the proposed annotation approach achieved notable speed-ups compared to traditional sample-by-sample annotation.

Although Fallgren et al.’s paper \cite{fallgren_interspeech} offers an important demonstration of how 2DV‑based GUIs can accelerate annotation of acoustic data, our study addresses a substantially different problem setting: First, we focus on two distinct biomedical time‑series data modalities, both involving annotation tasks that are considerably more complex than binary speech activity detection. Second, our experiments evaluate how three different sample selection strategies generalize to these biomedical domains using modern self‑supervised learning embeddings rather than a single sample selection strategy on hand‑crafted acoustic features. Third, instead of focusing on labeling accuracy, we compare class distributions and downstream classification performance, which are central to the clinical utility of ML models. Fourth, in contrast to bulk labeling through region painting and label propagation via downsampling, we target labeling a small proportion of the data using point‑wise annotation, enabling controlled comparison of multiple sample selection strategies under a fixed annotation budget. Fifth, our study includes twelve annotators across two domains as opposed to a single annotator in one domain, allowing the assessment of method robustness across varying levels of annotator expertise. And lastly, in our approach all data points are visualized without downsampling and we do not perform label propagation, ensuring that annotation decisions are based solely on explicit human judgments. Together, these design choices position our work as an investigation of sample selection methods, including 2DV, for challenging biomedical downstream tasks, rather than an exploration of rapid bulk labeling in a single acoustic domain.

\begin{figure*}[t]
    \centering

    \begin{subfigure}{0.99\textwidth}
        \centering
        \includegraphics[width=\textwidth]{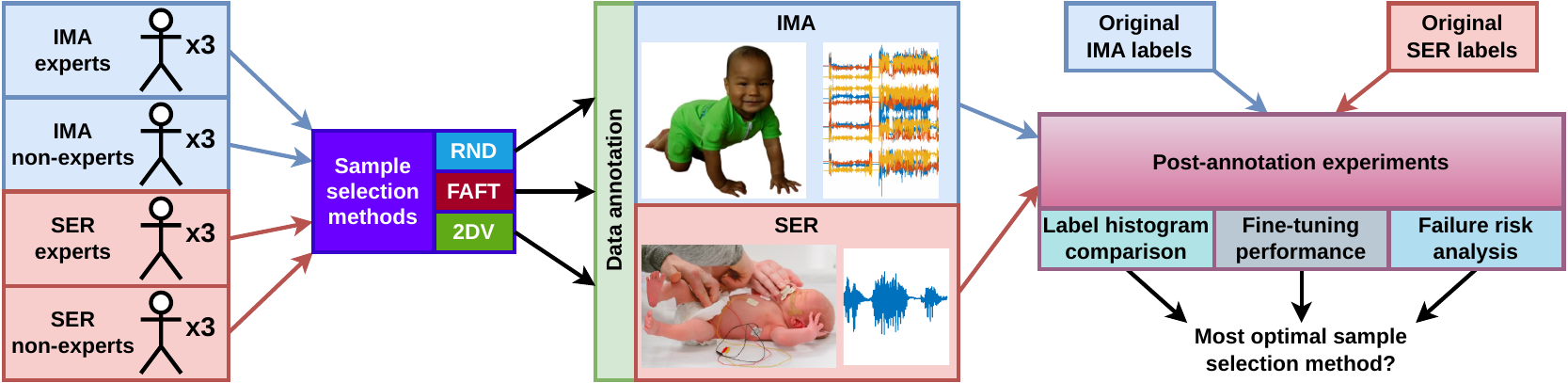}
        \caption{A block diagram of the overall study design.}
        \label{fig:tsexplorer_poc_study_block_diagram}
    \end{subfigure}

    \vspace{10pt} 

    \begin{subfigure}{0.74\textwidth}
        \centering
        \includegraphics[width=\textwidth]{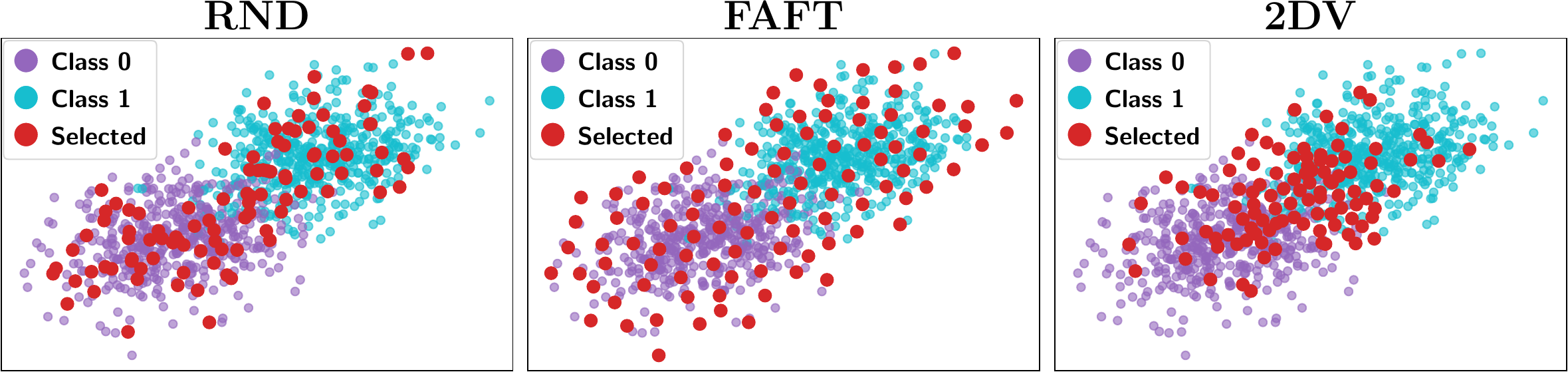}
        \caption{A 2D example of RND (left), FAFT (middle), and 2DV (right) sample selection strategies using simulated data, consisting of two classes with an overlapping region. In this example reflecting an idealized use case, the samples selected using the 2DV method are concentrated near the decision boundary between the two classes, thereby prioritizing samples whose annotation has high discriminative relevance. In contrast, both RND and FAFT distribute their sample selections throughout the feature space, including regions with minimal influence on the resulting decision boundary.}
        \label{fig:annotation_method_comparison_example_figure_simulated_data}
    \end{subfigure}
    \hspace{10pt}
    \begin{subfigure}{0.20\textwidth}
        \centering
        \includegraphics[width=\textwidth]{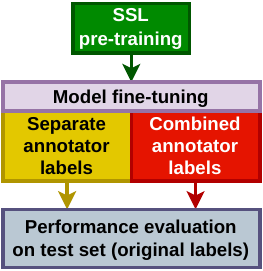}
        \caption{An illustration of the model performance evaluation procedure.}
        \label{fig:tsexplorer_poc_study_performance_evaluation_block_diagram}
    \end{subfigure}

    \caption{An overview of the study. \ref{fig:tsexplorer_poc_study_block_diagram}: Overall study design. \ref{fig:annotation_method_comparison_example_figure_simulated_data}: An example of different sampling strategies. \ref{fig:tsexplorer_poc_study_performance_evaluation_block_diagram}: Model performance evaluation procedure.}
    \label{fig:three_panel_layout}
\end{figure*}

\section{Sample selection methods} \label{sec_sample_selection_methods}

In our experiments, samples were selected for annotation using three different strategies: RND, FAFT (Section \ref{subsec_faft}), and 2DV-based data exploration (Section \ref{subsec_visualization}). RND served as a baseline method, in which a subset of samples was chosen uniformly at random (without replacement) from the dataset. FAFT, on the other hand, served as a representative method for an efficient algorithmic exploration of the feature space. However, as the decision boundaries between categories exhibited in various annotation tasks typically occupy certain manifolds of the feature space that are not uniformly distributed, we posited that both the RND and FAFT approaches allocate annotation effort into ``irrelevant'' parts of the dataset: The most relevant samples to annotate would lie close to the decision boundaries. Thus, our hypothesis was that the 2DV method would allow annotators to instinctively concentrate their effort on these areas (see Figure \ref{fig:annotation_method_comparison_example_figure_simulated_data} for an example).

To ensure fair method-wise comparison, no label propagation techniques (e.g. nearest-neighbor propagation) were applied in any of the selection strategies, even though such propagation could, in principle, be used with all the methods. This allows us to evaluate each strategy based solely on the samples it explicitly selects for annotation.

\subsection{Farthest-first traversal (FAFT)} \label{subsec_faft}

FAFT, also known as the $k$-center greedy algorithm, is an algorithm that selects samples by repeatedly choosing the farthest unannotated point from the current set of annotated points \cite{faft_original, faft_original_gonzalez}. Specifically, the algorithm first selects one sample at random and adds it to the set of annotated samples, $A$. Then, additional samples are added to $A$ one by one. The distance from a sample, $\bm{x}$, to the set $A$ is defined as

\begin{equation} \label{eq_faft_distance}
    d(\bm{x},A) = \underset{\bm{y}\in A}{\textrm{min}} \ d(\bm{x},\bm{y}) \ ,
\end{equation}
where $d$ is the distance metric. At each step, FAFT selects the next sample, $\mathbf{x}^*$, according to
\begin{equation} \label{eq_faft_sample_selection}
    \mathbf{x}^* = \underset{\mathbf{x}\notin A}{\arg\max} \ d(\mathbf{x},A) \ ,
\end{equation}
i.e. the point whose nearest annotated neighbor is as far away as possible. Consequently, FAFT sample selection yields a set of points that are maximally spread out in the feature space in terms of $d$ \cite{faft_original_gonzalez}.

\subsection{2D visualization-based data exploration (2DV)} \label{subsec_visualization}

\begin{figure*}[t]
  \centering
  \includegraphics[width=0.99\textwidth]{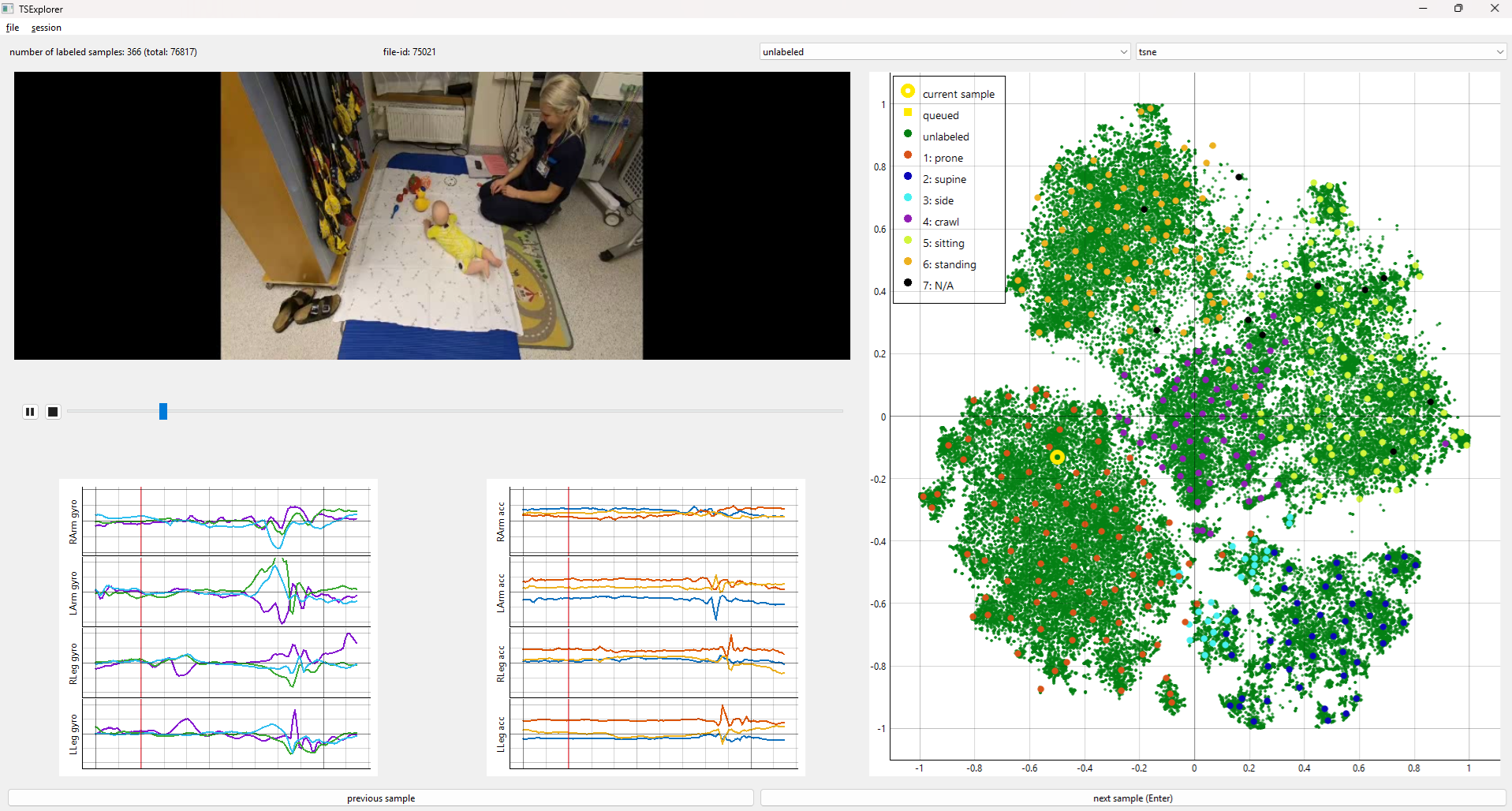}
  \caption{An example screenshot of the TSExplorer GUI as used for posture annotation for IMA in the present study. On the top left of the GUI, there is a video widget for video playback, and a scatter plot visualizing the entire dataset is located on the right. On the bottom left, there are two plots visualizing the accelerometer (right) and gyroscope (left) signals from the MAIJU-DS dataset. From top to bottom, the panels show the x, y, and z components for the right arm, left arm, right leg, and left leg. Users can select samples from the scatter plot for annotation in any order, and the annotation can be performed either via a drop-down menu or through keyboard shortcuts. The color of each data point in the scatter plot reflects its current label (green denoting unlabeled samples), and the active sample is highlighted with a large yellow ring. The color code as well as the keyboard shortcut for each class can be seen in the scatter plot legend.}
  \label{fig:tsexplorer_maiju_posture_example_image}
\end{figure*}

Sample selection using 2DV-based data exploration was conducted using the TSExplorer GUI, which presents the entire dataset as a 2D scatter plot in which each point corresponds to a single sample. Each sample is linked via global indices to the full dataset, from which selected views (e.g. corresponding video, audio, and signal waveform(s)) can be presented to the user. The projection can be interactively switched between t-SNE \cite{tsne}, PCA, and UMAP \cite{umap_arxiv, umap_nature_rev_met_primers} 2D embeddings, enabling the annotator to explore complementary low-dimensional representations of the underlying high-dimensional data. Samples can be selected for inspection or labeling in any desired order via left-clicking, while right-clicking enqueues samples for later review; queued samples can be browsed sequentially using the \textit{next sample} button, or by directly left-clicking them in the scatter plot. Users can annotate samples either using a drop-down menu, or by using keyboard shortcuts. Each point in the scatter plot is colored based on its current label (unlabeled samples are shown in green), and the active sample is highlighted with a large yellow ring. The interface supports continuous zooming, with point sizes adapting dynamically to the current zoom level. Furthermore, labels can be modified at any time, including previously annotated samples. Figure \ref{fig:tsexplorer_maiju_posture_example_image} shows an example screenshot of the TSExplorer GUI as used for data annotation in the present study for IMA. For a listing of TSExplorer features and an example screenshot of SER data annotation, see Appendix \ref{sec_appendix_tsexplorer_ui_description}.

\section{Experimental setup} \label{sec_experimental_setup}

Figure \ref{fig:tsexplorer_poc_study_block_diagram} provides an overview of the experimental setup. Twelve annotators (five female), consisting of three experts and three non‑experts for each dataset (IMA and SER; see Section \ref{subsec_data}), participated in the study. During the annotation process of the present study, annotators labeled data as if no prior labels existed, even though both datasets contained original labels (described in Section \ref{subsec_data}). Each annotator used all three sample selection methods (RND, FAFT, and 2DV), which were presented in a randomized order (Section \ref{subsec_data_annotation}). After the annotation process, we performed a series of post‑annotation analyses (Section \ref{subsec_post_annotation_experiments}) to compare the sample selection methods and determine the most optimal one. These analyses were primarily conducted using the labels obtained in the present study, while the original dataset labels were only used as external references (e.g. for label distribution comparisons and for obtaining reference performance levels). For all ML model training procedures, we used an NVIDIA Tesla V100 GPU to train our models, and the code was implemented using PyTorch version 2.5.0.

\subsection{Data} \label{subsec_data}

We conducted our experiments across two distinct biomedical time‑series data modalities, IMA and SER, both of which include two classification tasks and come with existing labels. Table \ref{table:annotation_method_comparison_data_info_table} summarizes the IMA and SER datasets, which are described in detail in Sections \ref{subsubsec_data_maiju} and \ref{subsubsec_data_nicu_a}, respectively. In the present study, these datasets serve two complementary roles: First, they provide the unlabeled samples that annotators label during our annotation experiments, i.e. annotators do not access or use the original labels. Second, their original labels serve as ground truth references in the post-annotation experiments, for example when evaluating downstream model performance.

\begin{table}[t]
    \centering
    \caption{A summary of the datasets (IMA: MAIJU-DS; SER: NICU-A) used in the present experiments.}
    \includegraphics[width=0.39\textwidth]{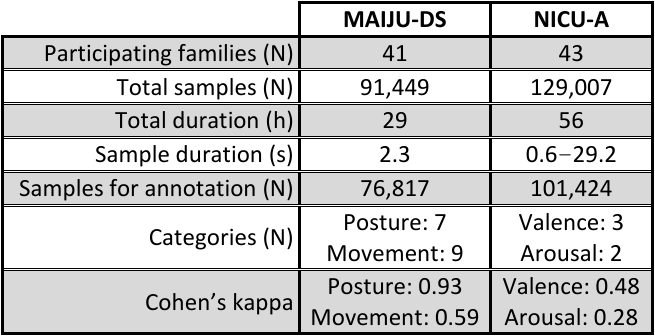}
    \label{table:annotation_method_comparison_data_info_table}
\end{table}

\subsubsection{Infant motility assessment} \label{subsubsec_data_maiju}

For IMA, we use a labeled dataset of multi-sensor IMU data collected with the MAIJU smart wearable \cite{maiju_commsmed}, referred to as MAIJU-DS in the present study. MAIJU is designed to provide an objective and scalable method for quantifying infants' posture and movement during natural play, addressing the clinical need for reliable and privacy-preserving tools for early monitoring of infant motor development. Traditional approaches based on direct expert observation are often subjective, resource-intensive, and compromise ecological validity \cite{maiju_commsmed}, whereas parental questionnaires provide only limited and often noisy sources of information \cite{maiju_pediatrics}. Because early motor patterns reflect the integrity of developing neural systems, objective characterization of spontaneous movements can support earlier identification of atypical development and provide complementary information for clinical decision-making \cite{maiju_commsmed}.

MAIJU-DS includes 41 recordings and distinct participants (age range 4--18 months; three participants with diagnosed motor disabilities) with a total of approximately 29 h of annotated IMU data. Each recording contains parallel video footage that was used to support data annotation. The recordings were obtained either in a semi-controlled setup at the infants' homes ($N=17$ recordings) or in a home-like laboratory environment ($N=24$), and they capture spontaneous, free-form infant play. The MAIJU wearable contains one IMU sensor at controlled locations on the proximal part of each limb, totaling four sensors. Each sensor records tri-axial accelerometer and tri-axial gyroscope signals at a sampling frequency of 52~Hz, resulting in 24-channel time-series data (four limbs, three axes, accelerometer and gyroscope). Following the pre-processing steps of \cite{maiju_commsmed}, the raw signals were interpolated to a common 52-Hz time base, gyroscope bias was removed, and a temporal median filter was applied. The pre-processed signals were then split into 120-sample frames (approximately 2.3 s) with a 50\% overlap, yielding a total of 91,449 annotated frames for the 29-hour dataset (76,817 annotated frames after removing frames where the infant was out of screen or handled by caregiver).

MAIJU-DS contains frame-level annotations for two parallel tasks: posture classification and movement classification. The movement labels include nine categories: still, roll left, roll right, pivot left, pivot right, proto movement, elementary movement, fluent movement, and transition. The posture labels include seven categories: prone, supine, left side, right side, crawl posture, sitting, and standing. Each frame was annotated independently by two or three domain experts \cite{maiju_commsmed}, and all recordings were quality-screened during the annotation process to ensure the reliability of the labels. Difficulty-wise, the movement annotation task is considerably more challenging, as reflected by its lower inter-rater agreement rate (multiclass Cohen's kappa=0.59) compared to the posture annotation task (kappa=0.93).

\subsubsection{Speech emotion recognition} \label{subsubsec_data_nicu_a}

For our SER experiments, we use the Finnish subset of the FinEst NICU Audioset (NICU-A) \cite{nicu_a_speechcomm}, collected as part of the APPLE study \cite{apple_tutkimus} to characterize the early auditory environment of preterm infants in NICUs. Caregiver speech constitutes a major source of early social and auditory input for preterm infants \cite{apple_tutkimus, anette_paperi, anette_uudempi_paperi}, and its emotional tone may influence later neurodevelopment. Therefore, the objective evaluation of the sound environment, including the affective quality of speech, is clinically relevant for understanding early developmental risk factors and for informing family‑centered care practices in the NICU \cite{pipari_study, preterm_birth_associated_with_depression, anette_paperi, anette_uudempi_paperi}.

The Finnish NICU‑A subset was recorded at the NICU of Turku University Hospital using LENA recorders \cite{lena_orig}. The recorders were placed near the infant to capture naturalistic caregiver and staff speech, and participants were informed to keep the device near the infant at all times. Families were eligible to participate in the study if the infant was born before 32 gestational weeks and did not have acute life‑threatening diseases or major congenital anomalies. Recordings were made when the infants reached a postmenstrual age of approximately 32 weeks, typically in relatively calm single‑family rooms where parents served as the primary talkers, with clinical staff occasionally present.

The dataset consists of 43 continuous 16-h recordings (16-kHz sampling frequency), totaling 688 hours of audio. Using LENA’s broad‑class diarization, each recording was segmented into utterances and assigned to speaker groups. After removing utterances shorter than 600 ms, the dataset includes 129,007 utterances (average duration 1.57 s, approximately 56 h of speech).

To construct training and evaluation sets, 35 families were assigned to the training set and eight families to the gold standard (GS) test set, ensuring that both sets were clinically and environmentally representative (e.g. infant health status and respiratory support). A subset of the training set was selected for annotation using an AL-based sampling strategy \cite{nicu_a_speechcomm}, with two annotators labeling disjoint subsets for emotional valence (negative/neutral/positive) and arousal (high/low) resulting in a total of 5198 labeled training set utterances. For the GS test set, three domain experts independently annotated the same samples from a randomly selected subset of the GS test set utterances, with access to 10 s of preceding audio context. Majority voting was then used to determine final labels, and samples without a consensus were removed, resulting in 345 labeled GS test set utterances. In terms of annotation difficulty, both tasks can be described as challenging (GS set kappa scores of 0.48 and 0.28 for valence and arousal, respectively).

\subsection{Data annotation} \label{subsec_data_annotation}

The annotation process was carried out separately for the IMA and SER datasets (MAIJU-DS and NICU-A, respectively), both of which included two annotation tracks (IMA: posture and movement; SER: valence and arousal). For each dataset, three expert and three non-expert annotators participated, resulting in six annotators per dataset and 12 annotators altogether (five female). For all annotation tasks, expert annotators were defined as individuals with previous annotation experience for the respective task. Accordingly, annotators without such experience were categorized as non-experts.

All annotators used each of the three sample selection methods (RND, FAFT, and 2DV) for data annotation. For every method, annotators first produced $N$ labels for the first track (posture or valence), followed by another $N$ labels for the second track (movement or arousal). We used the same annotation categories as in MAIJU-DS and NICU-A for data annotation, with the exception of merging laterality-specific categories in IMA. For example, the categories \texttt{roll left} and \texttt{roll right} were merged into a single \texttt{roll} category. Consequently, for the post-annotation experiments (Section \ref{subsec_post_annotation_experiments}), the laterality-specific categories in MAIJU-DS were also merged for harmonization. It was only possible to assign a single category to each sample, and, if no category was applicable (e.g. a non-speech sound in SER annotation), annotators could also label a sample as being erroneous. These erroneous samples were then removed from further analyses.

The value of $N$ was 360 for IMA and 400 for SER, so each annotator generated 2160 labels in total for IMA or 2400 labels for SER across the three methods. The slightly higher $N$ for SER was chosen to account for the higher number of erroneous samples (such as noisy samples with unrecognizable speech) observed in the NICU-A dataset in order to have approximately the same number of labeled samples for both datasets after erroneous samples were removed. For IMA, one sample-to-be-annotated corresponds to one frame (approximately 2.3 seconds) of multi-sensor IMU data (video, accelerometer, and gyroscope data) from the MAIJU-DS dataset. For SER, one sample corresponds to an utterance (audio, mean 1.57 s, median 1.16 s, range 0.6--29.2 s) from the NICU-A training set.

The ordering of sampling methods in the annotation tasks was determined using a constrained randomization procedure: The 2DV method was assigned to appear once in each position (first, second, and third) across the three annotators in each group (expert or non-expert). For each annotator, the remaining two methods (RND and FAFT) were then assigned randomly to the remaining positions. This ensured a balanced placement of the visualization method while maintaining randomness for the other two methods.

For the RND and FAFT sample selection methods, the TSExplorer GUI was used without the 2D scatter plot (as in the 2DV method), enabling a similar annotation interface between the sampling methods. For all sample selection methods, the user could see the total number of samples and the current number of labeled samples during the entire annotation process, and samples were annotated one at a time to enable a fair comparison between the methods. For the 2DV method, annotators were given complete freedom in sample selection, i.e. they could choose samples in any order using any of the available 2D projections, without suggested guidelines or annotation strategies. A different random seed was used for each annotator for RND and FAFT in both annotation tracks to ensure that the annotators did not annotate the same sets of samples. For a listing of the TSExplorer features as used for data annotation and example screenshots of the GUI, see Section \ref{subsec_visualization} and Appendix \ref{sec_appendix_tsexplorer_ui_description}.

For both FAFT and 2DV, we used self-supervised learning (SSL)-based features obtained from models pre-trained with the algorithm by Vaaras et al. \cite{pfml}. We utilized SSL-based representations because they eliminate the need for hand-crafted features while capturing informative structure in the data, and they have shown potential in AL and sample selection \cite{interpretability_driven_sample_selection_ssl, vaaras22_interspeech, issl_al, ssl_class_balanced_al, density_based_al_ssl}. Model pre-training was carried out separately for both IMA and SER following the IMA and SER experiments in Vaaras et al. \cite{pfml}, with the only modification of using smaller model variants with three (instead of six) Transformer encoder blocks. For IMA, we used the 160-dimensional outputs of the CNN-based sensor encoder as the SSL features, whereas for SER we used the 128-dimensional Transformer output features (see \cite{pfml} for further details). For FAFT, cosine distance on the SSL features was used as the distance metric. For 2DV, the 2D projections (t-SNE, PCA, and UMAP) were computed from the SSL features, resulting in approximately 77,000 and 101,000 visible data points for IMA and SER, respectively.

In terms of computational costs associated with the current approach, SSL model pre-training, SSL-based feature extraction using the pre-trained model, and 2DV computation (t-SNE, PCA, and UMAP) were performed as offline pre-processing steps before data annotation, after which the resulting features were provided as inputs to TSExplorer. For example, for IMA, model pre-training required approximately 19 hours, SSL-based feature extraction using the pre-trained models took 20 seconds (both run on an NVIDIA Tesla V100 GPU), and computing the 2DVs took 7 minutes on a single CPU core (3.9 GHz). Note that all these steps can be conducted before the actual annotation process, and do not thereby affect the runtime performance of TSExplorer.

\subsection{Post-annotation experiments} \label{subsec_post_annotation_experiments}

After obtaining all annotations, we conducted a set of post-annotation experiments to compare the three data sampling methods (RND, FAFT, and 2DV). These post-annotation experiments were \textit{label histogram comparison}, \textit{model fine-tuning performance}, and \textit{failure risk analysis}, which are described in Sections \ref{subsubsec_label_histogram_comparison}, \ref{subsubsec_ft_performance}, and \ref{subsubsec_failure_risk_analysis}, respectively.

\subsubsection{Label histogram comparison} \label{subsubsec_label_histogram_comparison}

To determine which sampling method yielded the most consistent label distributions and most effectively captured underrepresented class categories, we analyzed label histograms and their variability across different experimental conditions. For both annotation tracks in IMA (posture and movement) and SER (valence and arousal), and for each of the three annotator groups (experts, non‑experts, and all annotators), we generated label histograms and calculated their standard deviations (SDs) for all sampling methods under comparison. As a reference, we used histograms derived from the full MAIJU-DS dataset for IMA and from the GS subset of the NICU‑A dataset for SER.

\begin{figure*}
  \centering
  \includegraphics[width=0.99\textwidth]{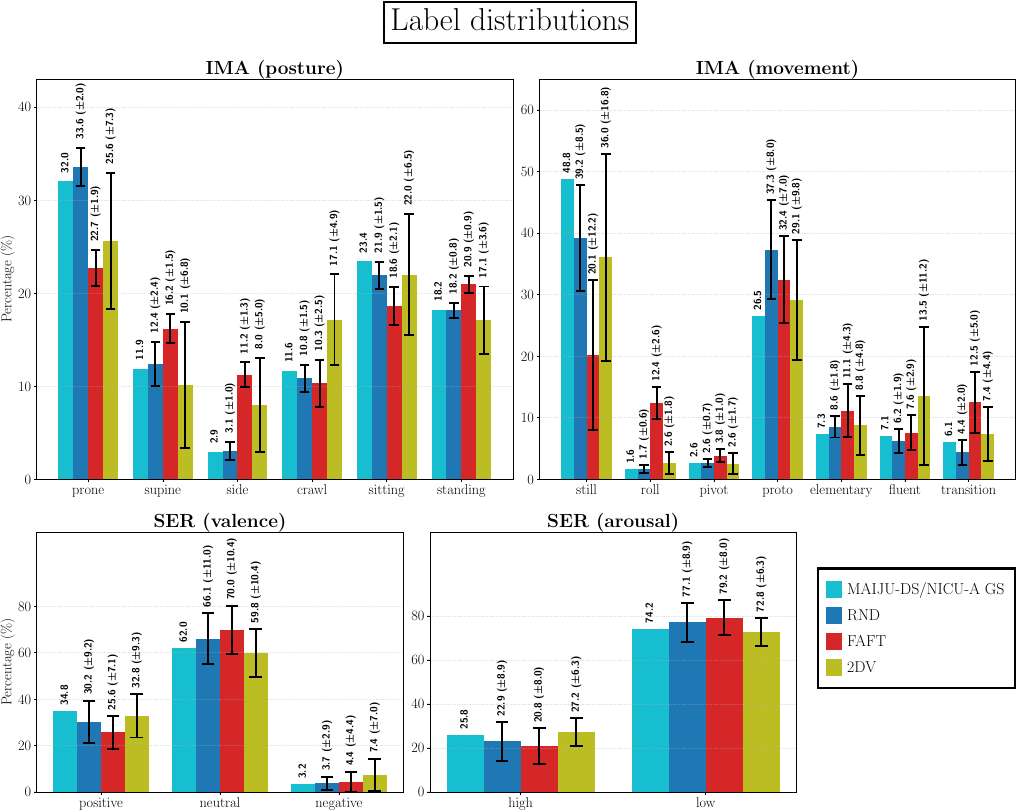}
  \caption{The combined label distributions (±SD) for all annotators in terms of IMA (top left: posture; top right: movement) and SER (bottom left: valence; bottom right: arousal). Each label-wise histogram group is organized from left to right in the following order: MAIJU-DS or NICU-A GS reference distribution, RND, FAFT, and 2DV.}
  \label{fig:label_distributions_all_annotators_only}
\end{figure*}

\subsubsection{Model fine-tuning performance} \label{subsubsec_ft_performance}

We evaluated model classification performance across different experimental conditions under two scenarios: \textit{separate annotator labels} and \textit{combined annotator labels}. In the separate‑label scenario, SSL pre-trained models (Section \ref{subsec_data_annotation}) were fine-tuned for classification using the labels from a single annotator within each sampling method, and the final result was computed as the average performance across all three annotators. In the combined‑label scenario, we merged the labels of either three annotators (experts or non‑experts) or all six annotators within a sampling method, and the aggregated labels were then used for fine-tuning. We used the majority label for samples with multiple labels, or, if no majority existed, we randomly selected one of the tied labels. For computing the test performance, the IMA test set consisted of all MAIJU-DS samples (with their corresponding labels) that were not annotated during the annotation process by any annotator, while for SER the test set was the GS set of NICU-A. In addition to training models using all available labels, we examined how model performance develops as the number of annotations accumulates by additionally running each fine-tuning experiment using the first 50, 100, ..., 250, and 300 labels from each annotator. Each fine-tuning experiment was run 10 times to mitigate the effect of randomness, and we report the mean performance across these runs. For a depiction of the model performance evaluation procedure, see Figure \ref{fig:tsexplorer_poc_study_performance_evaluation_block_diagram}.

Following Vaaras et al. \cite{pfml}, each model fine-tuning was conducted in two stages: First, we add two randomly initialized fully-connected Gaussian error linear unit layers followed by a softmax function after the Transformer network, and we fine-tune these layers separately while the weights of the rest of the model are frozen. Second, we fine-tune the entire model with a learning rate warm-up period. In contrast to Vaaras et al. \cite{pfml}, we use all training data for model fine-tuning without creating separate training and validation splits. Consequently, we train the models for 100 epochs in both fine-tuning stages, and we reduce the learning rate by a factor of 0.5 every 30 epochs instead of reducing it based on validation-loss plateauing. For further details on the fine-tuning process, refer to Section 4 of Vaaras et al. \cite{pfml}. Also, following the approach of Vaaras et al. in \cite{nicu_a_speechcomm} and \cite{pfml}, the valence categories \texttt{neutral} and \texttt{negative} were combined into a \texttt{non-positive} category to form a binary classification.

As model performance metrics, we use the unweighted average F1 score (UAF1) for IMA and the unweighted average recall (UAR) for SER. UAF1 is defined as the unweighted average of class-specific F1 scores, and it was selected for IMA both to account for the large class imbalances in MAIJU-DS, and also to enable comparison with prior work on MAIJU data (e.g. \cite{maiju_commsmed, iar_2_0, pfml}). UAR, defined as the unweighted average of class-specific recalls, was selected for SER as it is the most commonly used evaluation metric in SER literature, is also suitable for unbalanced classes, and enables comparison both with other SER studies and with previous studies on NICU-A (e.g. \cite{nicu_a_speechcomm, iar_2_0, pfml}).

As a reference for fine-tuning performance, we computed a \textit{topline} performance level for IMA by training the model using all MAIJU-DS samples that had been annotated at least once in our experiments, replacing the annotations with the corresponding labels from the MAIJU-DS dataset. For SER, we computed a \textit{reference} performance level by fine-tuning on the labeled subset of the NICU-A training set. Since this subset was selected using clustering-based active learning (\cite{nicu_a_speechcomm}), and therefore already incorporates a specific sampling strategy, it should be interpreted as a reference performance rather than a strict topline.

\subsubsection{Failure risk analysis} \label{subsubsec_failure_risk_analysis}

We conducted a failure risk analysis to evaluate the robustness of the sampling methods by examining three complementary aspects of risk. First, we quantified \textit{model performance failure} by marking a failure event whenever the model's performance fell below a relative threshold defined as 0.9 times the best performance obtained within the same configuration. In analyses involving individual annotators, the reference performance was the highest mean performance across annotators for each method. This measure reflects the likelihood that labels produced through a given sampling method lead to insufficient model accuracy. Second, we assessed \textit{rare class coverage failure} by determining whether each method successfully captured underrepresented categories. Rare classes were identified using the reference distributions of the MAIJU-DS dataset for IMA and the NICU‑A GS set for SER. In IMA, rare classes were those with less than 10\% representation in MAIJU-DS, corresponding to one posture class (side) and five movement classes (roll, pivot, elementary, fluent, transition). In SER, the rare categories were those with the lowest proportions in the NICU-A GS set, one for valence and one for arousal. A failure occurred if a method, or an individual annotator in the single‑annotator analyses, captured less than half of the proportional representation of the rarest class relative to its reference distribution. For example, if the rarest class accounted for 2\% of the reference data, a failure was assigned when the method produced less than 1\% representation. This criterion directly evaluates the risk that a method fails to capture important but infrequent categories. Third, we examined \textit{label distribution instability} by quantifying the divergence between annotators' label distributions as the mean pairwise Hellinger distance between annotators' label histograms. This measure captures the sensitivity of label distributions to annotator‑specific factors, and, since annotator expertise and backgrounds may vary widely in real clinical settings, methods with high instability pose a greater risk.

For the failure risk analysis, we only include the classification results from models trained with all available annotations. Because the 2DV method allowed unrestricted sample selection with a relatively tight annotation budget, early annotation stages often resulted in uneven coverage of the feature space (see Section \ref{subsec_separate_annotators_results}). This, on the other hand, can distort comparisons at smaller annotation counts. Therefore, using all available labels ensures a fair risk analysis across the sample selection methods.

We summarized the overall risk level of each sampling method by computing a combined failure risk score for every condition. This score aggregates the method-specific ranks (1, 2, or 3) assigned for model performance failure, rare class coverage failure, and label distribution instability. For a sampling method $m$ in condition $c$, the combined score is defined as
\begin{equation}
    S(m, c) = \sum_{k \in \mathcal{M}(c)} R_k(m, c),
\end{equation}
where $\mathcal{M}(c)$ contains all applicable failure risk metrics (i.e. model performance failure, rare class coverage failure, and label distribution instability) for the given condition, and $R_k(m, c)$ is the rank assigned to method $m$ under metric $k$. Lower values of $S(m, c)$ correspond to lower overall failure risk. Ranks were assigned using a dense ranking scheme, where methods with tied performance receive the same rank, and the next rank increases by one. Since we did not include fine-tuning models with the labels of annotator pairs (i.e. two annotators) in the present classification experiments, the model performance failure assessment is likewise omitted in the two-annotator cases.

\section{Results} \label{sec_results}

\subsection{Label histogram comparison results} \label{subsec_label_histogram_comparison_results}

Figure \ref{fig:label_distributions_all_annotators_only} presents the combined label distributions of all annotators. For detailed label distributions organized by annotator group, refer to Appendix \ref{sec_appendix_label_distributions_annotator_group}. For IMA, both FAFT and 2DV selected more samples from rare classes (relative to the MAIJU-DS label distribution) than RND, which expectedly was on a par with the baseline distribution. Regarding movement labels, FAFT notably emphasized selecting samples from the \texttt{roll} category while giving the most common category of MAIJU-DS, \texttt{still}, comparatively little attention. In contrast, 2DV provided a more balanced emphasis across the rare classes while still prioritizing them more effectively than RND. Here, it is essential to note that the FAFT sampling strategy, which prioritizes selecting samples that are maximally novel within the feature space, does not necessarily align with maximizing relevance for the annotation task. In settings where balanced representation across classes is important, such novelty‑driven sampling may therefore be suboptimal.

For SER, 2DV was the most effective sampling method for selecting samples from the rarest category of the NICU-A GS set, \texttt{negative}, in terms of valence. For arousal, 2DV was also slightly more effective than the other sampling methods in selecting samples from the minority class, \texttt{high}.

\begin{figure*}
  \centering
  \includegraphics[width=0.7\textwidth]{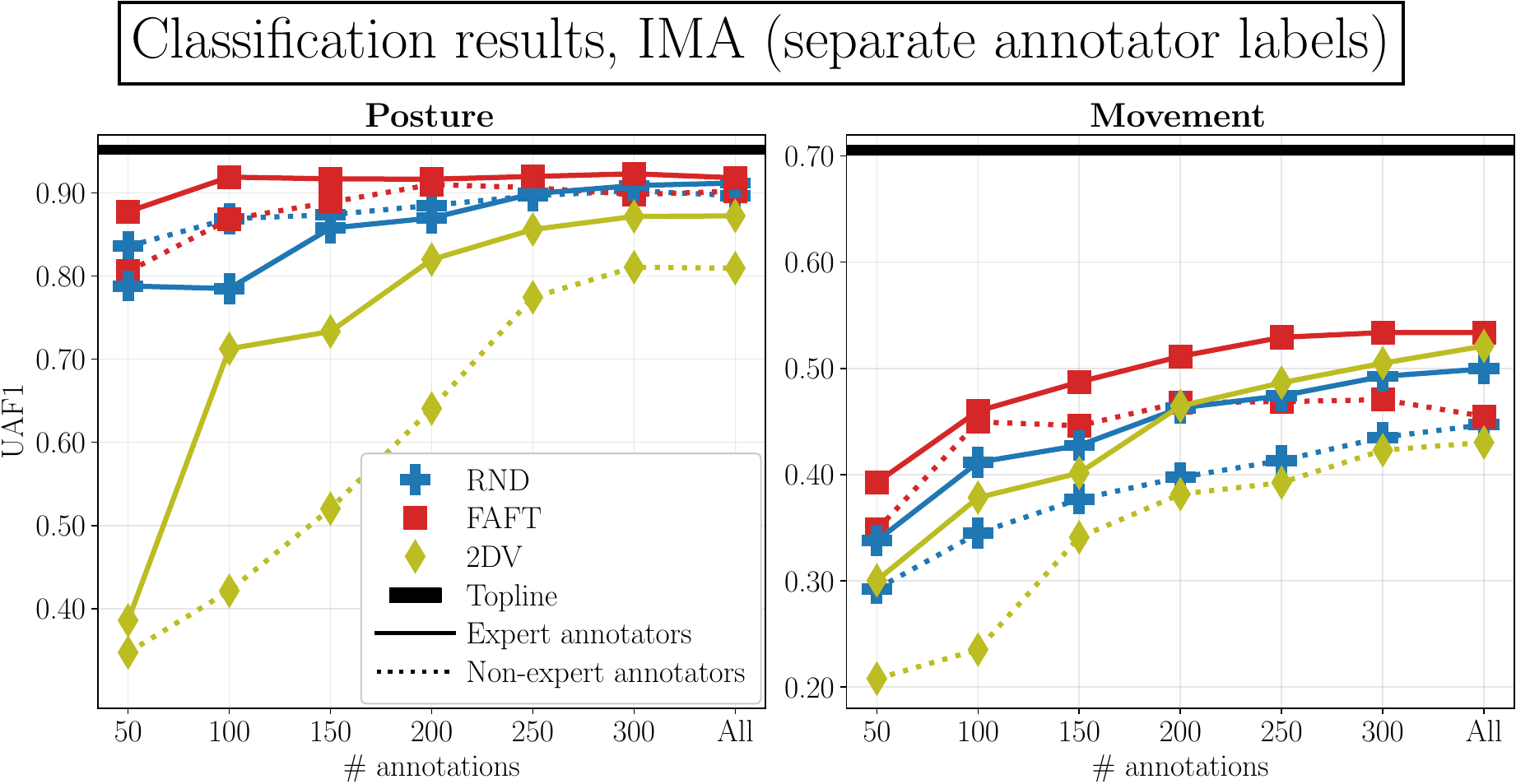}
  \caption{The mean classification results for IMA when training models with annotator-wise labels separately, organized by annotator group (solid line: expert annotators; dotted line: non-expert annotators) and by classification task (left: posture; right: movement). A topline performance level (black horizontal line), derived from a model trained using MAIJU-DS labels, is shown for comparison.}
  \label{fig:maiju_compare_finetuning_performance_individual_annotators}
\end{figure*}

\begin{figure*}
  \centering
  \includegraphics[width=0.7\textwidth]{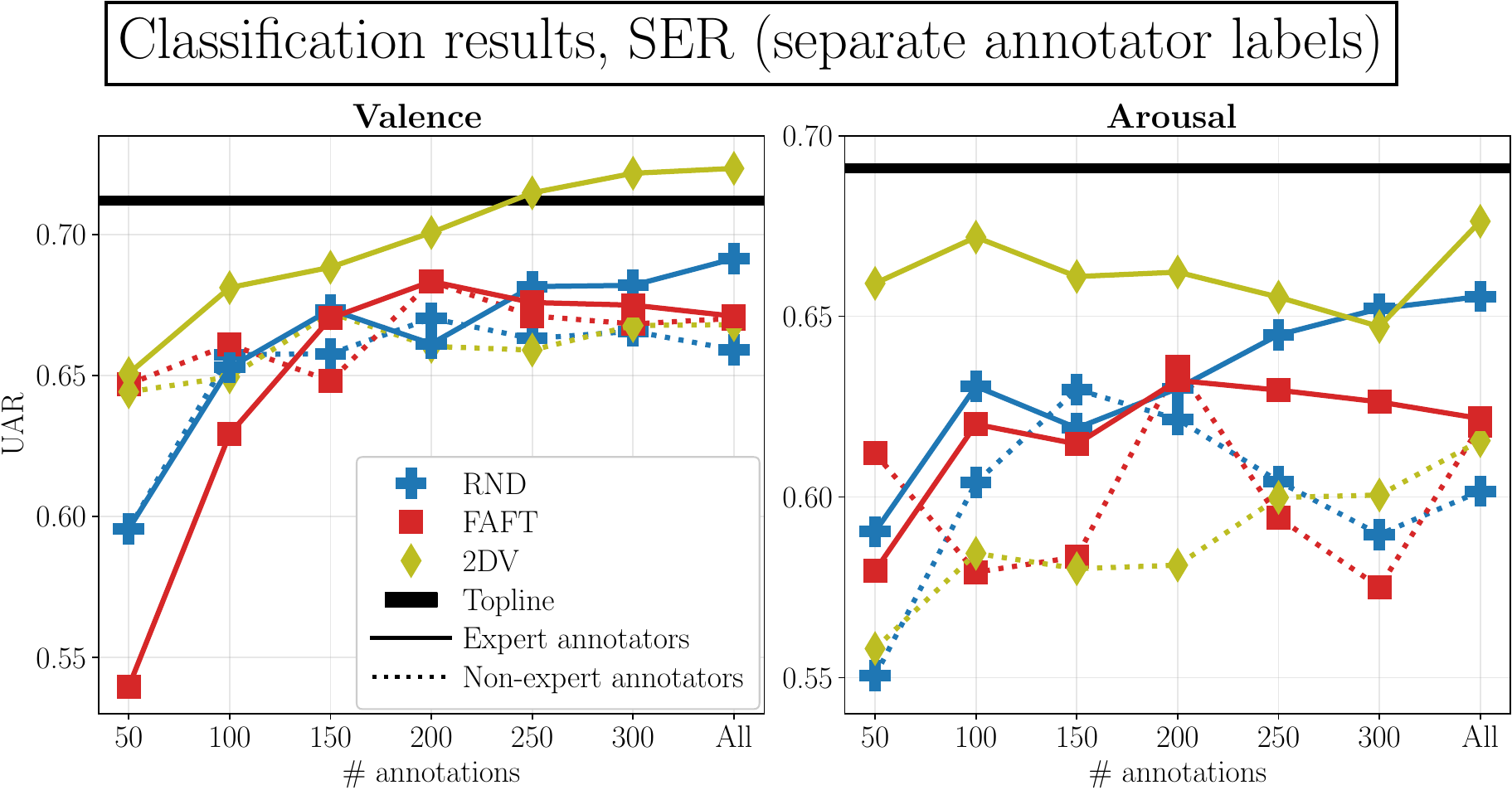}
  \caption{The mean classification results for SER when training models with annotator-wise labels separately, organized by annotator group (solid line: expert annotators; dotted line: non-expert annotators) and by classification task (left: valence; right: arousal). A reference performance level (black horizontal line), derived from a model trained using the labeled subset of the NICU-A training set, is shown for comparison.}
  \label{fig:nicu_compare_finetuning_performance_individual_annotators}
\end{figure*}

In IMA, RND produced the most consistent label distributions overall in terms of histogram variance. FAFT typically yielded slightly higher variability than RND, with the exception of the `expert' and `non-expert' posture conditions, where FAFT achieved the most consistent distributions (see Figure \ref{fig:maiju_label_distributions} in Appendix \ref{sec_appendix_label_distributions_annotator_group}). On the other hand, 2DV resulted in noticeably higher variability compared to the two other sampling methods in all cases, which is a natural consequence of the method: since we did not impose any restrictions on how annotators should select samples when using 2DV, annotators could follow diverse sample selection strategies. For example, one annotator may select samples from a small region within a single 2D projection, while another may attempt to select samples more evenly across all 2D projections. When further analyzing annotator-level label distributions, we indeed observed that some annotators focused more on selecting samples from specific 2D projection regions, leaving some categories sparsely represented or even entirely unlabeled, which suggests that given a higher annotation budget the results might have been more consistent (see the online appendix\footnote{\url{https://github.com/SPEECHCOG/tsexplorer_article_online_appendix}.} for 2DVs of the annotations). Such heterogeneous sample selection behavior may largely account for the increased variability in annotator-level histograms for 2DV compared to the RND and FAFT methods.

In SER, conversely, 2DV generally resulted in the lowest variability across annotators' label distributions. Two exceptions emerged in valence annotation (see Figure \ref{fig:nicu_label_distributions} in Appendix \ref{sec_appendix_label_distributions_annotator_group}): in the `all annotators' and `non-expert' groups, 2DV showed the highest variability, followed by FAFT, while RND produced the most consistent distributions. Outside these cases, the relative ordering of FAFT and RND varied depending on the specific condition. Compared to IMA results, the smaller variability observed in SER for 2DV may be related to the smaller number of class categories, which inherently limits the magnitude of distributional divergence across annotators.

\subsection{Classification results for separate annotator labels} \label{subsec_separate_annotators_results}

The classification performance results for training models with separate annotator labels are shown in Figures \ref{fig:maiju_compare_finetuning_performance_individual_annotators} and \ref{fig:nicu_compare_finetuning_performance_individual_annotators} for IMA and SER, respectively. For IMA, FAFT yielded the highest performance overall, followed by RND, whereas the 2DV method performed the worst by a large margin especially in posture classification. This likely originates from the unrestricted nature of the 2DV annotation procedure described in Section \ref{subsec_label_histogram_comparison_results}, where annotators may insufficiently sample categories or regions of the feature space within the constraints of the given annotation budget. This, in turn, hinders the model's ability to learn reliable class boundaries. The performance gap between 2DV and the other sampling methods was particularly noticeable when the number of training samples was small. A potential explanation is that, because annotators were given complete freedom in selecting samples with the 2DV method, even those who explored the 2D projections relatively broadly likely concentrated on specific regions of the feature space one at a time while progressing towards the target number of annotations. As a result, substantial portions of the 2D projections may remain unexamined during the early stages of annotation, leading to imbalanced coverage of the underlying data distribution and weaker model performance. However, as the number of annotations increases, the performance of the 2DV method gradually approaches the performance levels of FAFT and RND.

\begin{figure*}
  \centering
  \includegraphics[width=0.99\textwidth]{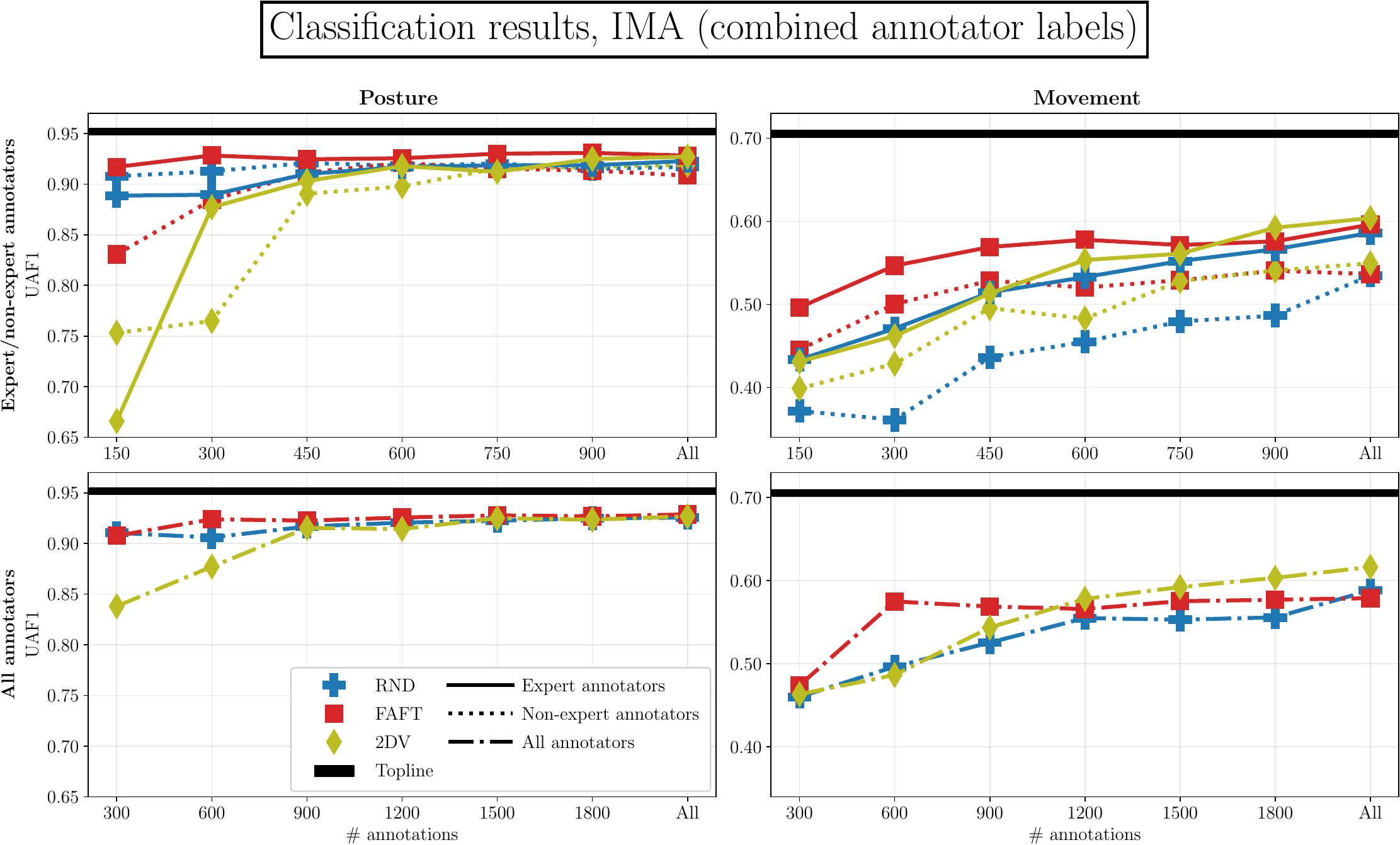}
  \caption{The classification results for IMA when training models with the labels of either three annotators (top row: expert and non-expert annotators) or six annotators (bottom row: all annotators), organized by classification task (left column: posture; right column: movement). A topline performance level (black horizontal line), derived from a model trained using MAIJU-DS labels, is shown for comparison.}
  \label{fig:maiju_compare_finetuning_performance}
\end{figure*}

\begin{figure*}
  \centering
  \includegraphics[width=0.99\textwidth]{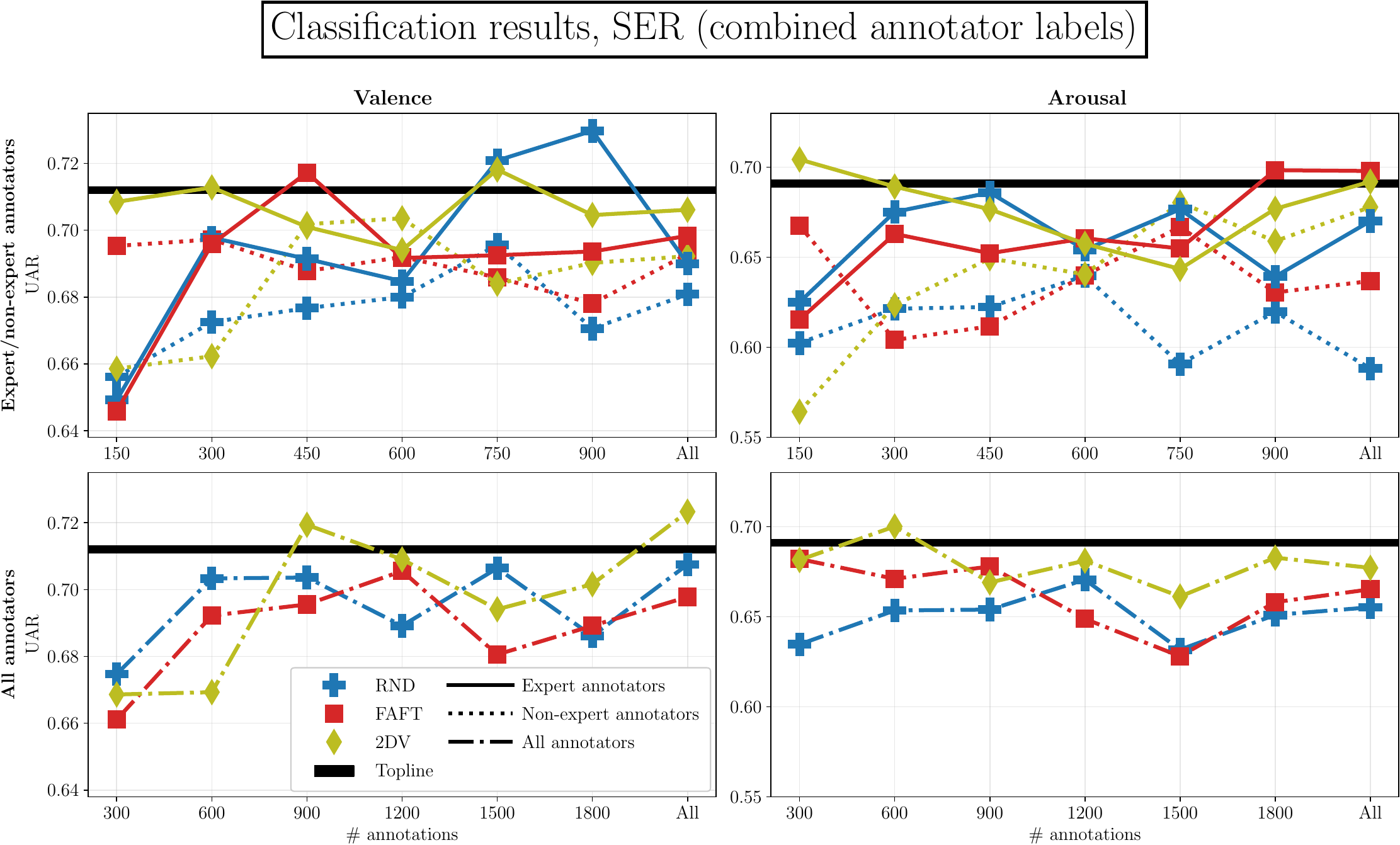}
  \caption{The classification results for SER when training models with the labels of either three annotators (top row: expert and non-expert annotators) or six annotators (bottom row: all annotators), organized by classification task (left column: valence; right column: arousal). A reference performance level (black horizontal line), derived from a model trained using the labeled subset of the NICU-A training set, is shown for comparison.}
  \label{fig:nicu_compare_finetuning_performance}
\end{figure*}

In contrast, for SER, the 2DV method excelled with expert annotators, outperforming both FAFT and RND by a substantial margin. This may be partly due to the structure of the SER annotation tasks, which consisted of either binary (arousal) or effectively binary (valence, due to class merging) classifications. Under such conditions, the free-form annotation style of the 2DV approach makes it less likely that annotators accidentally miss entire categories during the annotation process. With non-expert annotators, however, all annotation methods performed relatively equally. Here, it is important to note that the SER test set (NICU‑A GS) contains only 345 samples, which inherently increases the variance of classification results compared to those of IMA (over 70,000 test samples).

\subsection{Classification results for combined annotator labels} \label{subsec_combined_annotators_results}

Figures \ref{fig:maiju_compare_finetuning_performance} and \ref{fig:nicu_compare_finetuning_performance} show the classification performance obtained using combined annotator labels for IMA and SER, respectively. Regarding IMA, on average, the 2DV method generally outperformed RND and FAFT in cases with more than 200 annotations per annotator. With 200 or fewer annotations per annotator, FAFT achieved the best performance, likely because it provides more even coverage of the high-dimensional feature space when the number of annotations is small. Notably, the 2DV method surpassed FAFT with 250 or more annotations per annotator, which is still a modest annotation effort. For posture classification (i.e. the least challenging annotation task in terms of inter‑rater agreement), all sampling methods yielded similar performance when at least 150 annotations per annotator were available, but, with a smaller number of annotations, 2DV performed worst, which is in line with the sampling behavior discussed previously (Section \ref{subsec_separate_annotators_results}). Comparing with the results obtained using individual annotators' labels (Section \ref{subsec_separate_annotators_results}), it is evident that 2DV benefits substantially from combining heterogeneously sampled sets of labels. This effect is particularly noticeable for the more challenging IMA task, the classification of movement, where the combined labels produced the best overall performance, presumably because the aggregated annotations provide the classifier with the most diverse and relevantly sampled training set.

In terms of SER, the results contained more variance compared to the IMA results, as expected due to the small SER test set. Nevertheless, when comparing the area under each performance curve (higher is better), the 2DV method achieved the highest overall performance. With the labels of non-expert annotators only and a very small number of labeled samples per annotator (50 or 100), FAFT outperformed 2DV, showing a similar trend as with the IMA results.

Notably, in both the separate annotator labels (Section \ref{subsec_separate_annotators_results}) and combined annotator labels training scenarios, models trained using expert annotator labels consistently achieved higher classification performance than those trained with non-expert annotator labels when all available labels were used for model training. This finding highlights the importance of prior task-specific annotation experience in achieving higher-quality training labels.

\subsection{Failure risk analysis results} \label{subsec_failure_risk_analysis_results}

\begin{table*}
\centering
\caption{Combined failure risk score ranking for IMA (top) and SER (bottom). Lower scores indicate lower estimated risk, and each individual score is summed across classification tasks and different failure risk metrics. Metric abbreviations: \emph{cov} = rare class coverage failures; \emph{mod} = model performance failures; \emph{dis} = label distribution instability.}
\begin{tabular}{ccccc}
\toprule
Task & Annotator group & \# Annotators & Metrics used & Ranked methods (combined risk score) \\
\midrule
IMA & Expert & 1 & cov, mod, dis & FAFT (7.0) $=$ RND (7.0) $>$ 2DV (13.0) \\
IMA & Expert & 2 & cov, dis & FAFT (5.0) $=$ RND (5.0) $>$ 2DV (8.0) \\
IMA & Expert & 3 & cov, mod, dis & FAFT (7.0) $=$ RND (7.0) $>$ 2DV (10.0) \\
IMA & Non-expert & 1 & cov, mod, dis & FAFT (7.0) $>$ RND (8.0) $>$ 2DV (14.0) \\
IMA & Non-expert & 2 & cov, dis & FAFT (5.0) $=$ RND (5.0) $>$ 2DV (9.0) \\
IMA & Non-expert & 3 & cov, mod, dis & FAFT (7.0) $=$ RND (7.0) $>$ 2DV (10.0) \\
IMA & All & 6 & cov, mod, dis & RND (6.0) $>$ FAFT (8.0) $>$ 2DV (10.0) \\
\midrule
SER & Expert & 1 & cov, mod, dis & 2DV (6.0) $>$ RND (8.0) $>$ FAFT (11.0) \\
SER & Expert & 2 & cov, dis & 2DV (6.0) $>$ FAFT (8.0) $>$ RND (10.0) \\
SER & Expert & 3 & cov, mod, dis & 2DV (6.0) $>$ FAFT (10.0) $=$ RND (10.0) \\
SER & Non-expert & 1 & cov, mod, dis & RND (8.0) $>$ 2DV (10.0) $=$ FAFT (10.0) \\
SER & Non-expert & 2 & cov, dis & 2DV (8.0) $=$ FAFT (8.0) $=$ RND (8.0) \\
SER & Non-expert & 3 & cov, mod, dis & 2DV (8.0) $=$ FAFT (8.0) $>$ RND (9.0) \\
SER & All & 6 & cov, mod, dis & RND (7.0) $>$ 2DV (8.0) $>$ FAFT (9.0) \\
\bottomrule
\end{tabular}
\label{table:combined_failure_risk_ranking}
\end{table*}

Table \ref{table:combined_failure_risk_ranking} summarizes the failure risk analysis for IMA and SER, using the task‑summed combined failure risk scores described in Section \ref{subsec_post_annotation_experiments}. For IMA, which has the annotation tasks with the most categories with an imbalanced distribution, FAFT and RND achieved the lowest summed failure risk scores in all settings, whereas 2DV consistently showed distinctly higher risk. FAFT yielded the lowest overall risk in all cases except in the six‑annotator setting with all annotators combined, where RND achieved the lowest summed risk. Here, it is important to note that the heterogeneous sample selection behavior of the 2DV method (Section \ref{subsec_label_histogram_comparison_results}) accounted for a large proportion of the higher risk scores due to both rare class coverage failures and greater label distribution instability.

For SER, the results exhibited stronger dependence on annotator expertise: Among expert annotators, 2DV consistently achieved the lowest summed failure‑risk scores across one, two, and three annotators. FAFT and RND followed, with the exact ordering depending on the number of annotators. In contrast, for non‑experts with a single annotator, RND provided the lowest risk, while FAFT and 2DV showed higher but similar combined scores. With two or three non‑expert annotators, all methods produced either identical or nearly identical summed scores, indicating similar levels of robustness. In the six‑annotator setting, RND achieved the lowest summed failure-risk score, followed by 2DV and then FAFT.

Overall, FAFT and RND are generally the safest sampling methods in IMA with a limited labeling budget, regardless of annotator expertise. In SER, 2DV is the most robust choice in most situations, i.e. for expert annotators and cases with two or three non-expert annotators, whereas RND is more reliable for single-annotator cases with non-expert annotators and in the six-annotator condition. In general, RND is the safest choice in terms of overall risk, as indicated by the lowest total risk score computed across all tasks and conditions (totaling 105.0), followed by FAFT (110.0) and 2DV (126.0). In addition, RND stood out in six-annotator cases, suggesting a lower risk particularly for multi-annotator cases with potentially mixed levels of expertise.

\section{Discussion and conclusion} \label{sec_conclusion}

In the present study, we compared three sample selection methods for biomedical time-series data annotation using real human annotators: RND, FAFT, and a GUI-based approach utilizing 2DVs with the TSExplorer software. We conducted data annotation using 12 human annotators for two data modalities, with two tasks per modality: IMA using multi-sensor IMU data and SER using speech data. After the annotation process, we conducted a set of experiments to compare the sampling methods. Our findings show that 2DV most effectively selected samples from rare classes, but also exhibited greater label histogram variability across annotators in IMA due to the limited annotation budget and the free-form annotation style of the method. In several IMA cases, consisting of multi-class annotations, annotators sparsely labeled or completely omitted some classes, which negatively impacted average classification performance in IMA when models were trained separately using annotator-specific labels. Under these single-annotator conditions, FAFT achieved the best classification performance in IMA. For SER, consisting of either binary or ternary annotation tasks, however, 2DV generally produced low label distribution variability and excelled in single-annotator classification performance among expert annotators, while also being on a par with FAFT and RND among non-expert annotators. When combining labels from multiple annotators for model training, already a modest annotation effort was sufficient for the 2DV method to outperform the two other sampling methods in both IMA and SER. This result highlights the advantage of combining heterogeneously sampled sets of labels, particularly when individual annotators follow diverse data exploration strategies. We also analyzed the riskiness of each sampling method and found that RND was generally the safest option when annotator expertise or the number of available annotators is uncertain, whereas 2DV had the highest average risk due to its greater label distribution variability, mostly caused by the limited annotation budget. Overall, the results demonstrate that 2DV offers the greatest potential among the tested sampling methods, in particular when annotations from multiple annotators can be combined and/or the labeling budget can be expanded. Albeit, 2DV contains a higher risk of suboptimal outcomes under tightly constrained annotation scenarios, emphasizing the importance of careful deployment choices.

It is important to note that many of the 2DV-related failure modes observed in this study arise from interaction design choices rather than from inherent limitations of 2DV-based annotation. In the present experiments, annotators using the 2DV method were deliberately given unrestricted freedom in sample selection to enable a more naturalistic assessment of the method and to identify potential limitations. In practical deployment scenarios, relatively lightweight GUI constraints or workflow instructions could help alleviate the risk of uneven class coverage while preserving the benefits of human-driven data exploration. Such design choices could include visual indicators of underexplored regions of the 2D space, feature space coverage targets, or explicit user instructions encouraging more systematic feature space exploration.

It should be emphasized that the present experiments were primarily designed as a proof-of-concept: both the number of annotators and the annotation budget per annotator were limited, and, consequently, these constraints should be taken into account when interpreting the results. For example, in the failure risk analysis (Section \ref{subsec_failure_risk_analysis_results}), larger annotation counts may lead to different results, such as reducing the rare class coverage failures observed in imbalanced multi-category tasks (IMA) for the 2DV method or narrowing the differences between the tested methods. Future studies should thus primarily target higher annotation budgets across diverse annotation tasks to better determine the specific applications of 2DV-based annotation. Another limitation concerns the choice of feature representations used to compute 2DVs in TSExplorer: In the present work, we relied on SSL-based features, which, as with any choice of feature representation, require domain knowledge of ML and additionally of the data modalities and annotation tasks (here, IMA or SER). Without such expertise, the visualization pipeline cannot be applied directly. In cases where prior ML knowledge is limited, for example, one could instead use more conventional signal-based features derived from the literature and compute 2DVs from those representations. As an example, in SER, such features could include log-mel spectrograms or mel-frequency cepstral coefficients. Nevertheless, without integrating any domain knowledge at all into the TSExplorer pipeline, there is no guarantee that the resulting 2DVs are sensible in terms of a given class categorization.

Across both IMA and SER, labels from expert annotators consistently resulted in superior model performance compared to labels from non-expert annotators when all available labels were included in model training. This finding suggests that when additional annotation resources become available, prioritizing task-relevant expertise may yield greater benefits than increasing the number of annotations from non-expert annotators, particularly in single-annotator cases. Nevertheless, the 2DV method benefited strongly from combining annotations across multiple annotators regardless of annotator expertise, indicating that annotation aggregation from multiple annotators and annotator expertise play complementary roles.

Potential directions for future work include experimenting with interactive 3D data projections, which may offer improved data explorability compared to using 2DVs. Additionally, TSExplorer currently supports only categorical labels, whereas extending the framework to handle continuous-valued labels would increase its applicability. Moreover, integrating feature extraction into the TSExplorer pipeline is another promising enhancement. A further direction for potential future work is to iteratively update the 2DVs based on the accumulating annotations. For example, after every $N$ annotated samples, a feature extraction model could be retrained using the newly obtained labels, and a new 2DV could be generated based on the updated features. Implementing such an adaptive workflow would require mechanisms to prevent inappropriate model updates. For instance, if annotations are concentrated in a small region of the feature space and several classes remain underrepresented, the resulting updated feature space and its 2DVs might not be optimal. 

Additionally, in the present experiments, annotators were provided with three complementary 2DVs of the data, namely t-SNE, PCA, and UMAP, to highlight different aspects of the underlying high-dimensional data and to reduce reliance on a single 2D projection method. At any given time, only one 2DV was visible, and annotators were free to switch between 2DVs without restrictions. Consequently, individual annotators differed notably in how they utilized the available visualizations, with some relying primarily on a single projection while others explored multiple 2DVs during the annotation process. While different 2DV algorithms can influence annotator behavior, the dominant effects observed in the results of our experiments were driven by annotation budget, annotator expertise, and task-specific class structure rather than by the particular choice of 2D projections. However, a systematic evaluation of 2D projection-specific impact on the results would require additional experiments with constraints on annotator interaction.

Although user experience was not an explicit component of the present experiments, it is noteworthy that the common sentiment among annotators was that the 2DV approach made the annotation task more engaging. In post‑experiment interviews, annotators described 2DV as making the annotation task more interesting, enjoyable, and purpose‑driven. Such factors, while often overlooked in annotation studies, may help mitigate intra‑annotator effects such as cognitive fatigue arising from overly monotonous or highly variable sample streams. These observations suggest that beyond its quantitative performance, 2DV may offer motivational and cognitive benefits that support sustained annotation quality. However, these aspects should be examined more systematically in future work.

Even though 2DV offers benefits such as annotator engagement and improved selection of samples from rare classes, the classification performance gains observed under limited annotation budgets are most noticeable when heterogeneous sets of annotations from multiple annotators can be combined. When label aggregation or annotation guidance is not feasible, the IMA results of the present study suggest that algorithmic strategies, such as FAFT, remain a more robust baseline in single-annotator settings, particularly for multi-class classification tasks with notable class imbalances. In settings where annotator expertise or annotation behavior is highly uncertain, RND provides the most conservative choice, yielding the lowest overall failure risk across tasks and conditions. Yet, it is likely to require more annotation effort to reach the same level of classifier performance as the studied alternatives in several of the evaluated scenarios.

Interestingly, beyond the expert/non-expert distinction defined in this study, we also observed differences associated with annotators' prior experience in machine learning: Annotators with such experience tended to utilize the three available 2D projections more diversely, thereby covering a wider range of the corresponding feature spaces. Conversely, annotators without previous ML experience frequently concentrated their sample selections within a single 2D projection or localized regions within a projection. These findings suggest that annotators without previous ML experience may benefit from explicit instructions encouraging systematic exploration across all projections and throughout the 2D feature space. However, since we did not perform a systematic evaluation of annotators with and without previous ML experience, nor was annotator recruitment balanced with respect to this factor across modalities and the expert/non-expert distinction, further research is needed to better analyze this phenomenon. Furthermore, many of the issues we observed with the 2DV method were related to the heterogeneous sampling strategies adopted by different annotators. These issues may likewise be mitigated by providing explicit guidance on how to conduct the annotation process. In conclusion, the present results and our experience with the 2DV method show its great promise as a strategy we plan to apply for our own upcoming studies that require annotation of biomedical time-series data.

\printcredits

\section*{Code availability}

The annotation software that was used in the present experiments, TSExplorer, is publicly available online\footnote{\url{https://github.com/SPEECHCOG/TSExplorer}}.

\section*{Ethics statement}

This study did not involve the collection of new research data from human participants or animals. All human-related datasets analyzed were derived from previously approved studies, with ethical clearance and consent procedures reported in the original publications. The manual annotations used in this work were produced by contracted research staff as part of their employment duties and under the original informed consents and ethics approvals associated with the original data.

\section*{Declaration of competing interest}

The authors declare that they have no known competing financial interests or personal relationships that could have appeared to influence the work reported in this paper.

\section*{Acknowledgment}

The authors would like to express their gratitude to all annotators who participated in this study, and in particular to Santeri Heiskanen, who developed the initial version of TSExplorer. This work was supported in part by the Research Council of Finland (grants no. 343498 and 371243) and in part by the Sigrid Jusélius Foundation. The authors would like to thank Tampere Center for Scientific Computing for the computational resources used in this study. The author Einari Vaaras would also like to thank the Finnish Foundation for Technology Promotion for the encouragement grants and the Nokia Foundation for the Nokia Scholarship.

\section*{Appendices}
\appendix
\section{Key features of the data annotation platform} \label{sec_appendix_tsexplorer_ui_description}

\begin{figure*}[t]
  \centering
  \includegraphics[width=0.99\textwidth]{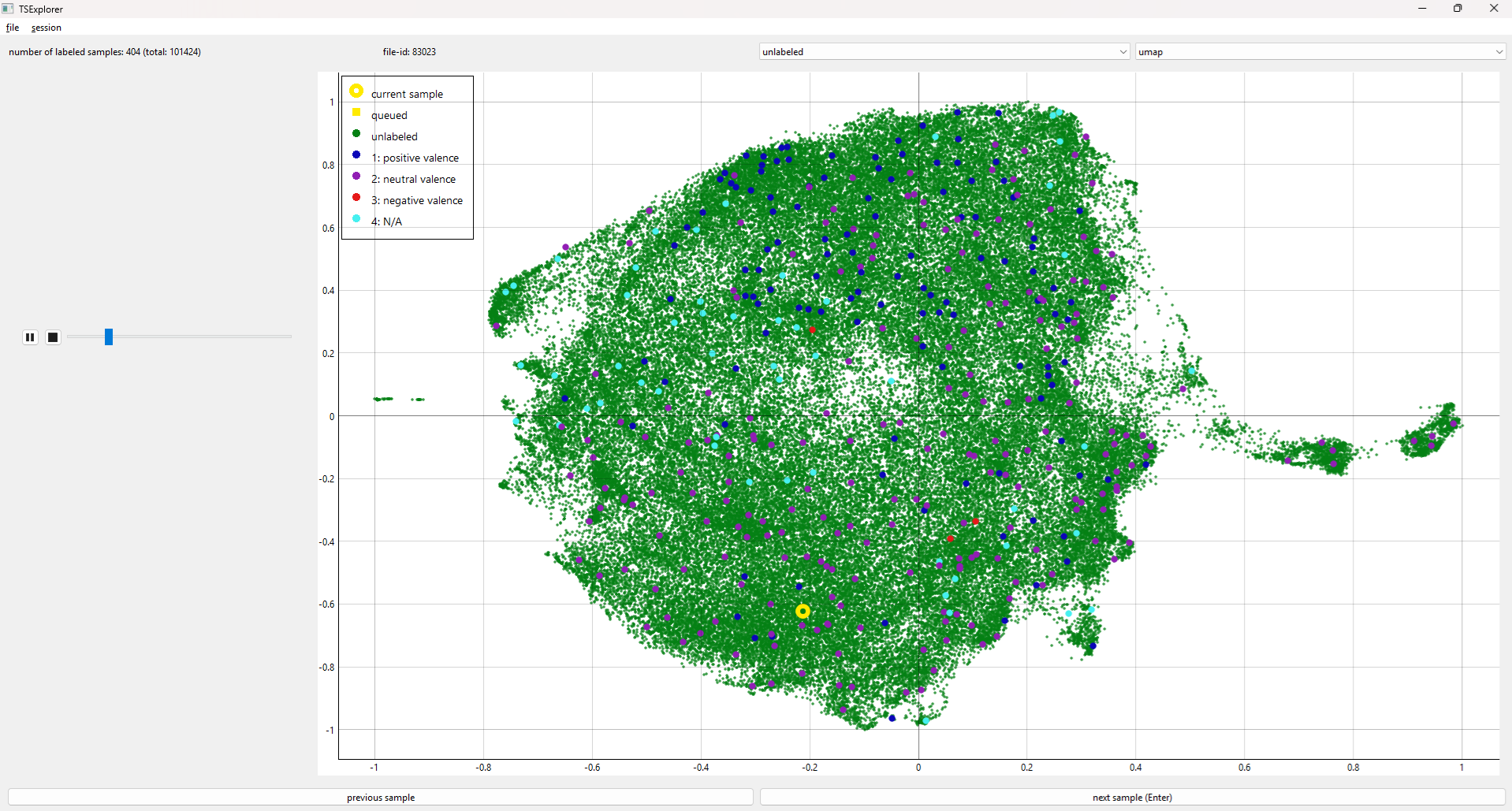}
  \caption{An example screenshot of the TSExplorer GUI as used for valence annotation for SER in the present study. On the left of the GUI, there is an audio widget used for audio playback, and a scatter plot visualizing the entire dataset is located on the right.}
  \label{fig:tsexplorer_nicu_valence_example_image}
\end{figure*}

Figures \ref{fig:tsexplorer_maiju_posture_example_image} and \ref{fig:tsexplorer_nicu_valence_example_image} show example screenshots of the TSExplorer GUI as used for data annotation in the present study for IMA and SER, respectively. The following list summarizes the key features of the annotation platform that were available to annotators when employing the 2DV-based sample selection method:
\begin{enumerate}
    \item Users can save the current annotation session using \textit{file} $\rightarrow$ \textit{save session / save as}.
    \item Users can load a previous annotation session using \textit{session} $\rightarrow$ \textit{load previous session}.
    \item Users can export their annotations as a CSV file using \textit{session} $\rightarrow$ \textit{export annotations as .csv}.
    \item Users can see the total number of samples and the current number of annotated samples.
    \item Users can see the file ID of the currently selected sample.
    \item For cross-platform compatibility, TSExplorer uses the VLC Media Player to play audio and video. Common media player functions like play, pause, stop, and scrolling function normally. Video without audio was used in IMA annotation, and audio only was used in SER annotation.
    \item Signal visualizations, such as waveforms (single- or multi-channel) or spectrograms, can be set visible, and their panels show a vertical line that is in sync with the scroll marker of the VLC media player. In the present experiments, the multi-channel signals (as shown in Figure \ref{fig:tsexplorer_maiju_posture_example_image}) were visible in IMA annotation, whereas signals were not visible in SER annotation.
    \item Users can see a 2D scatter plot representing the entire dataset. Each data point represents one sample-to-be-annotated. In the present study, one data point corresponds to approximately 2.3 seconds of multi-sensor IMU data (video, accelerometer, and gyroscope data) in IMA, or an utterance (audio) in SER. The user can change the visualization algorithm between t-SNE, PCA, and UMAP at any given time using a drop-down menu. The 2D projections have been computed from high-dimensional data, each visualization algorithm highlighting different aspects of the data.
    \item Users can select samples in the scatter plot for annotation or inspection by clicking the left mouse button. Users can also enqueue samples by pressing the right mouse button, after which the user can go through the queued samples one by one either with the \textit{next sample} button (shortcut: \textit{Enter} key), or by left-clicking the queued samples. Queuing is useful e.g. in cases when a user wants to inspect where a sample or a set of samples is located in the feature space for some other visualization algorithm than the one that is currently selected.
    \item Users can annotate samples either using a drop-down menu, or by using keyboard shortcuts. Each data point in the scatter plot is colored based on its assigned label (green for unlabeled samples), and the currently selected sample is highlighted with a large yellow ring. The color code as well as the keyboard shortcut for each class can be seen in the scatter plot legend. For example, pressing the key \textit{2} on the keyboard will assign the currently selected sample to the second class category. Using the scroll wheel of the mouse, users can adjust the zoom level of the scatter plot, and the data points dynamically change size based on the current zoom level. The labels of any sample, including those already labeled, can be changed at any given time.
    \item Users can go sample-by-sample backwards in the order they have annotated using the \textit{previous sample} button.
\end{enumerate}

Note that for the RND and FAFT sample selection methods, the same GUI was used without the 2D scatter plot to enable a similar annotation interface between the sampling methods. In these cases, instead of the scatter plot, the GUI contained a list of the available class categories and their keyboard shortcuts.

\section{Label distributions by annotator group} \label{sec_appendix_label_distributions_annotator_group}

\begin{figure*}
  \centering
  \includegraphics[width=0.99\textwidth]{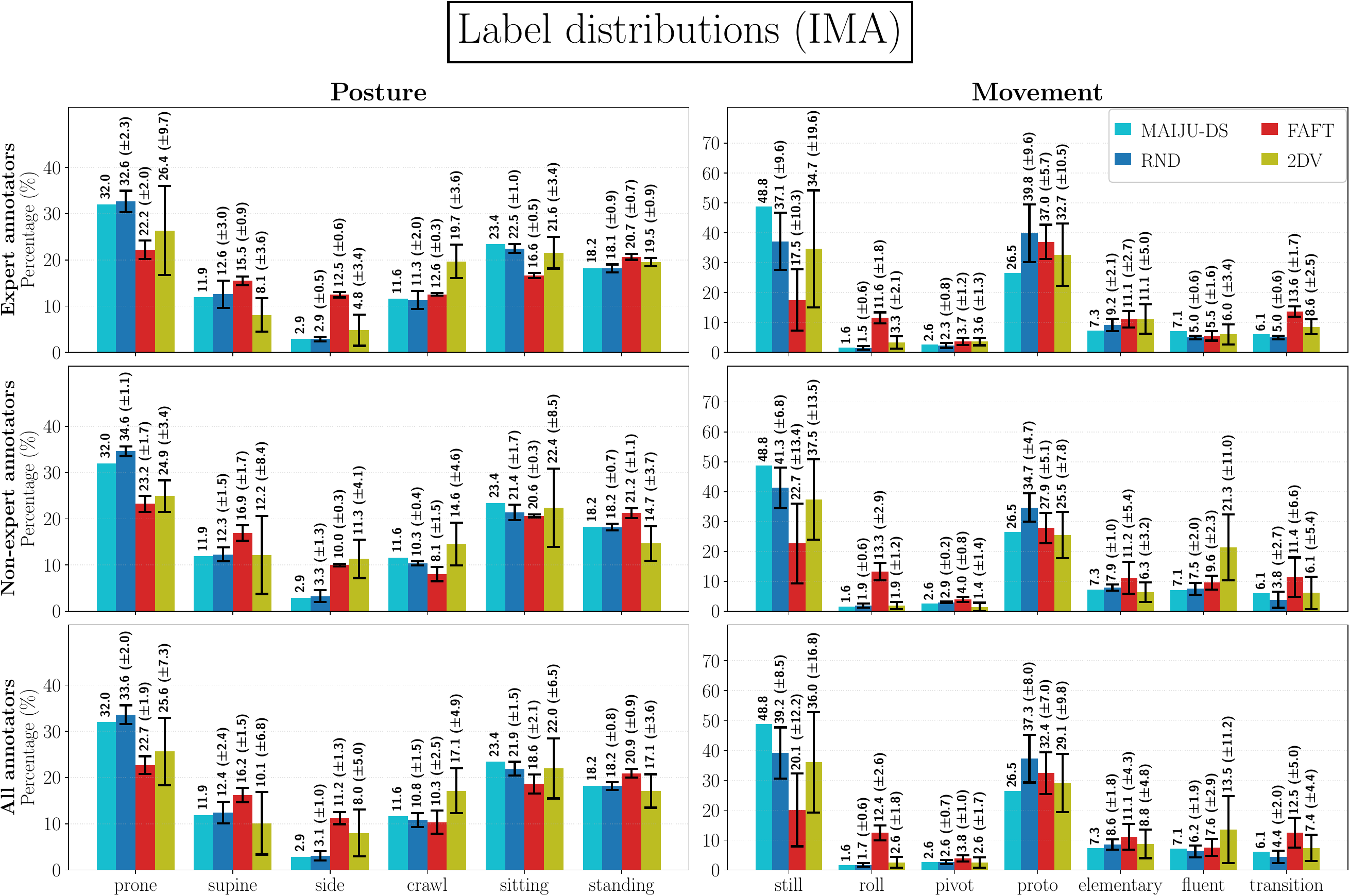}
  \caption{The combined IMA label distributions (±SD) for expert annotators (top row), non-expert annotators (middle row), and all annotators (bottom row) in terms of posture (left column) and movement (right column) labels. Each label-wise histogram group is organized from left to right in the following order: MAIJU-DS reference distribution, RND, FAFT, and 2DV.}
  \label{fig:maiju_label_distributions}
\end{figure*}

\begin{figure*}
  \centering
  \includegraphics[width=0.59\textwidth]{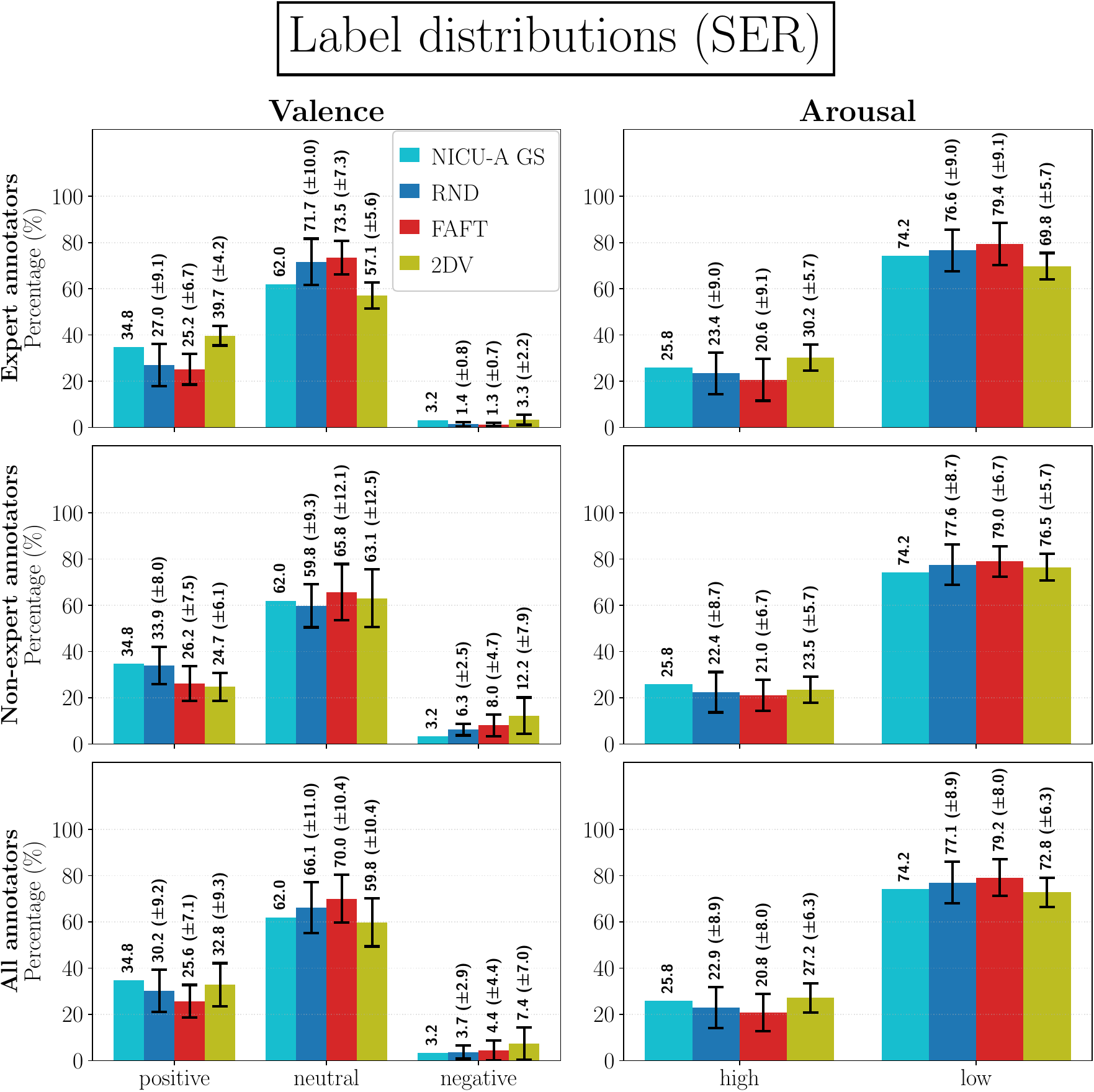}
  \caption{The combined SER label distributions (±SD) for expert annotators (top row), non-expert annotators (middle row), and all annotators (bottom row) in terms of valence (left column) and arousal (right column) labels. Each label-wise histogram group is organized from left to right in the following order: NICU-A GS set reference distribution, RND, FAFT, and 2DV.}
  \label{fig:nicu_label_distributions}
\end{figure*}

Figures \ref{fig:maiju_label_distributions} and \ref{fig:nicu_label_distributions} present the label distributions for IMA and SER, respectively. Both figures present the label distributions organized by annotator group.

\bibliographystyle{cas-model2-names}

\bibliography{cas-refs}



\end{document}